\documentclass[runningheads]{llncs}
\usepackage{graphicx}
\usepackage{amsmath,amssymb} 
\usepackage{color}
\usepackage[width=122mm,left=12mm,paperwidth=146mm,height=193mm,top=12mm,paperheight=217mm]{geometry}

\usepackage{leftidx}
\usepackage{bm}
\usepackage{subfig}
\usepackage{psfrag}
\usepackage{xcolor}
\usepackage{psfrag}
\usepackage{multicol}
\usepackage{hhline}

\usepackage[]{cite}
\usepackage{color}

\begin{document}
\title{A Minimal Closed-Form Solution for \\ Multi-Perspective Pose Estimation \\ using Points and Lines} 

\titlerunning{Multi-Perspective Pose Estimation using Points and Lines}
%
\author{Pedro Miraldo\textsuperscript{1}
\and Tiago Dias\textsuperscript{1}
\and Srikumar Ramalingam\inst{2}}
%
\authorrunning{P. Miraldo, T. Dias, and S. Ramalingam}
%

\institute{Instituto Superior T\'{e}cnico, Lisboa \\ 
\email{\{pedro.miraldo,tiagojdias\}@tecnico.ulisboa.pt}
\and
School of Computing, University of Utah\\
\email{srikumar@cs.utah.edu}}
\maketitle              

\begin{abstract}
We propose a minimal solution for pose estimation using both points and lines for a multi-perspective camera. In this paper, we treat the multi-perspective camera as a collection of rigidly attached perspective cameras. These type of imaging devices are useful for several computer vision applications that require a large coverage such as surveillance, self-driving cars, and motion-capture studios. While prior methods have considered the cases using solely points or lines, the hybrid case involving both points and lines has not been solved for multi-perspective cameras. We present the solutions for two cases. In the first case, we are given 2D to 3D correspondences for two points and one line. In the later case, we are given 2D to 3D correspondences for one point and two lines. We show that the solution for the case of two points and one line can be formulated as a fourth degree equation. This is interesting because we can get a closed-form solution and thereby achieve high computational efficiency. The later case involving two lines and one point can be mapped to an eighth degree equation. We show simulations and real experiments to demonstrate the advantages and benefits over existing methods. 
\end{abstract}      
\section{Introduction}
\label{sec:intro}
Pose estimation is a fundamental problem that is used in a wide variety of applications such as image-based localization (complementary to global positioning units that are prone to suffer from multi-path effects), augmented/virtual reality, surround-view and birds-eye view synthesis from a car-mounted multi-camera system, and telemanipulation of robotic arms. The basic idea of pose estimation is to recover the camera position and orientation with respect to some known 3D object in the world. We typically associate a world coordinate frame to the 3D object and pose estimation denotes the computation of the rigid transformation between world frame and the camera coordinate frame. The problem setting for pose estimation is generally the same. We are given the correspondences between 2D features (points or lines) and 3D features for a calibrated camera, and the goal is to compute the rigid transformation between the camera and the world. 

It is relatively easier to develop non-minimal solutions, which utilize more than the minimum number of correspondences, for pose estimation problems.
However, non-practitioners of multi-view geometry algorithms may ponder over the following questions.
Is it really necessary to develop a complex algorithm to use the minimum number of features?
What is the need for hybrid algorithms~\cite{ramalingam11,kuang2013} that utilize both point and line features?
In practice with noisy data with outliers, non-minimal solvers produce inferior results compared to minimal solvers.
For example, in a challenging pose estimation scenario involving crowded roads and dynamic obstacles, we face the problem of having a large number of incorrect feature correspondences.
Having the flexibility of using point or line correspondences improves the robustness of the algorithms.
While there has already been several solvers, three main factors can be used to distinguish one solver from another: minimal/non-minimal, features, and camera system.
For example, we could think about a pose estimation algorithm for a central camera system that is minimal and uses only point correspondences. Tab.~\ref{tab:minimal_solutions_SOTA} lists several minimal solvers for different types of camera systems and features. 

\begin{table}[t]
\caption{List of minimal pose problems, for perspective, multi-perspective, and general camera models, using both points and/or lines.}
\label{tab:minimal_solutions_SOTA}
\begin{center}
\scalebox{0.85}{\begin{tabular}{|r|c|c|c|c|}
\hline
{\bf Minimal Problem} & {\bf \#Points/Lines} & {\bf \#Solutions} & {\bf Closed-Form} & {\bf Papers} \\
\hline \hline
Persepective with points & 3/0 & 4 & Yes & \cite{haralick94,gao03,kneip11,ke17,wang18}
\\ 
\hline
Perspective with lines & 0/3 & 8 & No & \cite{dhome89,chen90} \\ 
\hline
Perspective with points and lines & 2/1 & 4 & Yes & \cite{ramalingam11} \\ 
\hline
Perspective with points and lines & 1/2 & 8 & No & \cite{ramalingam11} \\ 
\hline \hline
Multi-Perspective with points & 3/0 & 8 & No & \cite{gim15}\\
\hline
Multi-Perspective with lines & 0/3 & 8 & No & \cite{gim16} \\ 
\hline
Multi-Perspective with points and lines & 2/1 & 4 & Yes & Ours \\ 
\hline
Multi-Perspective with points and lines & 1/2 & 8 & No & Ours \\ 
\hline
\hline
General camera with points & 3/0 & 8 & No & \cite{chen04,nister04,miraldo14}\\
\hline
\end{tabular}}
\end{center}
\end{table}


{\it Minimal solvers for central cameras:~}
The minimal camera pose solver using 3 point correspondences gives up to four solutions and can be computed in closed-form~\cite{haralick94,gao03,kneip11,ke17,wang18}. On the other hand, using 3 line correspondences, we get 8 solutions~\cite{dhome89,chen90}, requiring the use of slower iterative methods. In~\cite{ramalingam11}, two mixed scenarios were considered: 1) 2 points and 1 line yielding 4 solutions in closed-form; and 2) 1 point and 2 lines yielding 8 solutions, requiring the use of iterative methods. Non-minimal solvers using both points and lines have also been studied~\cite{Ansar2003,Vakhitov2016}.

{\it Minimal solvers for generalized cameras:}
The general camera model~\cite{grossberg01,sturm04,miraldo11,Sturm_cameramodels} is represented by the individual association between unconstrained 3D projection rays and pixels, i.e. when projection rays may not intersect at a single 3D point in the world. This problem was addressed for the pose estimation using three 3D points and their 2D correspondences~\cite{chen04,nister04,miraldo14}. On the other hand, no solutions were yet proposed for the case of using 3D straight lines and their images, neither the case of using the combination of points and lines. There are non-minimal solvers using both points and lines~\cite{Schweighofer08,miraldo15}.

{\it Minimal solvers for multi-perspective cameras:}
We refer to multi-perspective camera as a system that models multiple perspective cameras that are rigidly attached with respect to each other. Examples include stereo cameras, multi-camera system mounted on a car for surround-view capture, etc. While multi-perspective camera systems are non-central and can be treated as generalized cameras, they are not completely unconstrained. In both perspective and multi-perspective systems, 3D lines project as 2D lines and lead to interpretation "planes". The minimal solvers for multi-perspective cameras have been addressed independently for points~\cite{gim15} and lines~\cite{gim16}. In this paper, we propose a novel solution for pose estimation for multi-perspective cameras using both points and lines. We are not aware of any prior work that solves this problem. Both 3D point and line correspondences provide two degrees of freedom (as shown in~\cite{haralick94,gao03,chen04,nister04,kneip11,miraldo14,gim15,ke17,wang18} for points and~\cite{dhome89,chen90,gim16} for lines). Since we have 6 DOF, to compute the camera pose we need at least three lines and/or points\footnote{In other cases, such as catadioptric cameras 3D lines project as curves in images which, in theory, it is possible to get the 3D line parameters from a single image~\cite{Caglioti05,Swaminathan08,Cameo14}, providing therefore more degrees of freedom to solve the pose problem.}.

The pose estimation has also been studied under other settings~\cite{Camposeco2018,bujnak11,micusik13,kukelova13,kuang14,wu15,kukelova11,sweeney15,Micusik2013,camposeco16,sweeney14,ventura14,haner15,albl15}. The main contributions of this paper are summarized below:
\begin{itemize}
\item We present two minimal pose estimation solvers for a multi-perspective camera system given 2D and 3D correspondences (Sec.~\ref{sec:our_contrib} and Fig.~\ref{fig:intro}): 
\begin{enumerate}
    \item Using 2 points and 1 line, we get 4 solutions in closed-form; and
    \item Using 1 point and 2 lines, we get 8 solutions.
\end{enumerate}
\item The proposed solvers using both points and lines produce comparable or superior results to the ones that employ solely points or lines (Sec.~\ref{sec:exp});
\item While most prior methods require iterative solutions (See Tab.~\ref{tab:minimal_solutions_SOTA}), 2 points and 1 line correspondences yield an efficient closed-form solver (very useful for real-world applications such as self-driving); and
\item We demonstrate a standalone SLAM system using the proposed solvers for large-scale reconstruction (Sec.~\ref{sec:exp}).
\end{itemize}
\section{Problem Statement}

Our goal is to solve the minimal pose estimation for multi-perspective cameras using both points and lines. First, we present the general pose problem using points or lines (Sec.~\ref{sec:notation_absolute_pose}) and, then we define the minimal case studied in this paper (Sec.~\ref{sec:minimal_absolute_pose}).

\subsection{Camera Pose using Points or Lines}
\label{sec:notation_absolute_pose}
To distinguish between features in the world and the camera coordinate system we use $\mathcal{W}$ and $\mathcal{C}$, respectively. The camera pose is given by the estimation of the rotation matrix $\mathbf{R}_{\mathcal{C}\mathcal{W}} \in\mathcal{SO}(3)$ and the translation vector $\mathbf{t}_{\mathcal{C}\mathcal{W}}\in\mathbb{R}^{3}$ that define the rigid transformation between the camera and world coordinate systems:
\begin{equation}
    \mathbf{T}_{\mathcal{C}\mathcal{W}}\in\mathbb{R}^{4\times 4} =
    \begin{bmatrix}
    \mathbf{R}_{\mathcal{C}\mathcal{W}} & \mathbf{t}_{\mathcal{C}\mathcal{W}} \\
    \mathbf{0}_{1,3} & 1
    \end{bmatrix}.
\end{equation}

A multi-perspective camera is seen as a collection of individual perspective cameras rigidly mounted with respect to each other. We use $\mathcal{C}_i$ to denote the features in the $i$\textsuperscript{th} perspective camera. The transformations between the perspective cameras and the global camera coordinate system are known, i.e. $\mathbf{T}_{\mathcal{C}_i\mathcal{C}}$ is known, for all $i$. Next, we define the pose for the multi-perspective system.

\begin{figure}[t]
    \centering
    \subfloat[Two points and one line]{
      \psfrag{p2}{\small$\mathbf{p}_2^{\mathcal{W}}$}
      \psfrag{d2}{\small$\mathbf{d}_2^{\mathcal{C}_2}$}
      \psfrag{p3}{\small$\mathbf{p}_3^{\mathcal{W}}$}
      \psfrag{d3}{\small$\mathbf{d}_3^{\mathcal{C}_3}$}
      \psfrag{l1}{\small$\mathbf{l}_1^{\mathcal{W}}$}
      \psfrag{Pi1}{\small$\bm{\Pi}_1^{\mathcal{C}_1}$}
      \psfrag{Rt}{\small$_{\mathcal{C}}^{\mathcal{W}}\mathbf{T}$}
      \psfrag{Rt1}{\small$_{\mathcal{C}_1}^{\mathcal{C}}\mathbf{T}$}
      \psfrag{Rt2}{\small$_{\mathcal{C}_2}^{\mathcal{C}}\mathbf{T}$}
      \psfrag{Rt3}{\small$_{\mathcal{C}_3}^{\mathcal{C}}\mathbf{T}$}
    \includegraphics[width=0.48\textwidth]{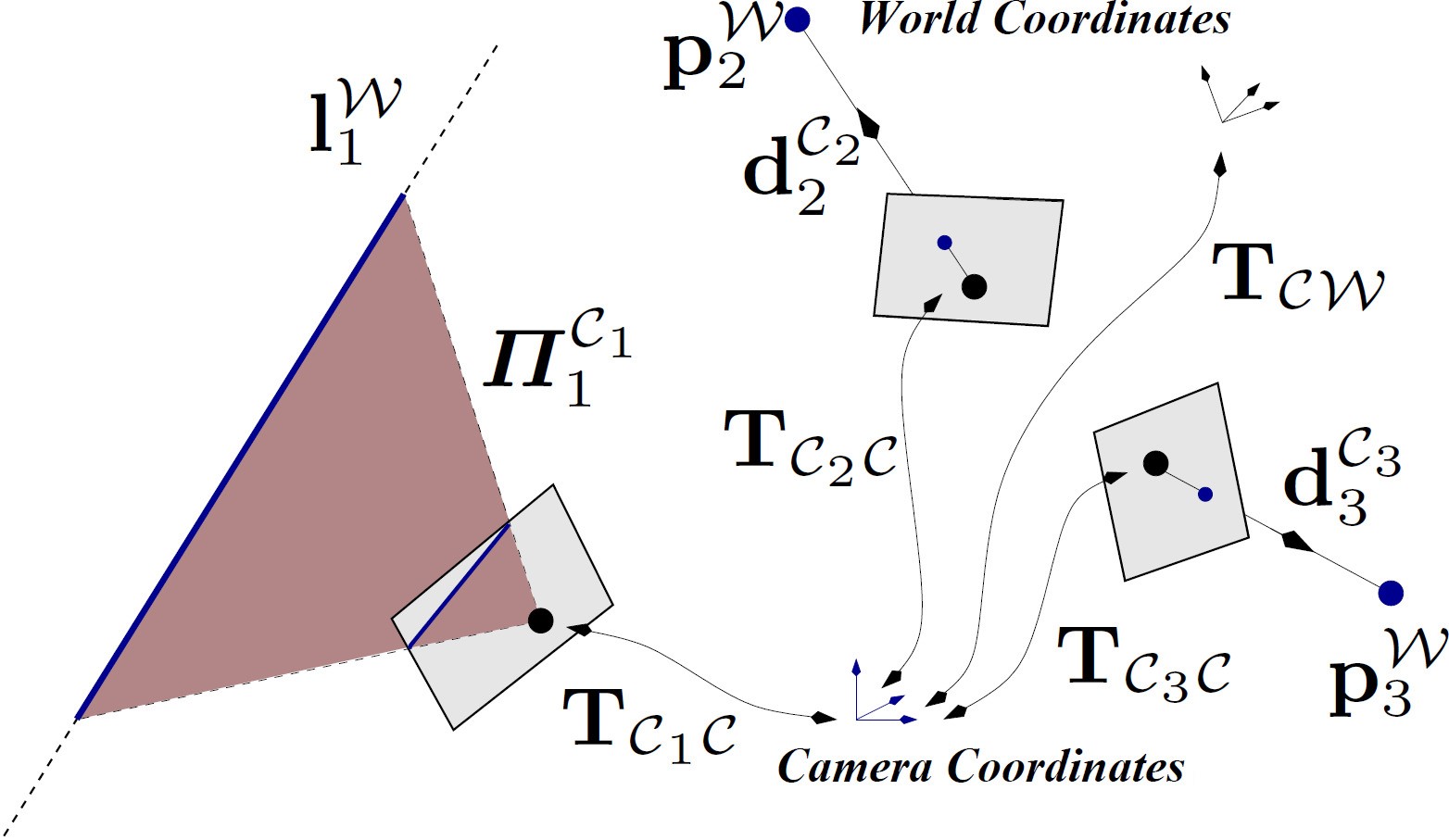}
    \label{fig:intro_2p1l}
    }
    \hfill 
    \subfloat[One point and two lines.]{
      \psfrag{p2}{\small$\mathbf{p}_2^{\mathcal{W}}$}
      \psfrag{d2}{\small$\mathbf{d}_2^{\mathcal{C}_2}$}
      \psfrag{l1}{\small$\mathbf{l}_1^{\mathcal{W}}$}
      \psfrag{Pi1}{\small$\bm{\Pi}_1^{\mathcal{C}_1}$}
      \psfrag{l3}{\small$\mathbf{l}_3^{\mathcal{W}}$}
      \psfrag{Pi3}{\small$\bm{\Pi}_3^{\mathcal{C}_3}$}
      \psfrag{Rt}{\small$_{\mathcal{C}}^{\mathcal{W}}\mathbf{T}$}
      \psfrag{Rt1}{\small$_{\mathcal{C}_1}^{\mathcal{C}}\mathbf{T}$}
      \psfrag{Rt2}{\small$_{\mathcal{C}_2}^{\mathcal{C}}\mathbf{T}$}
      \psfrag{Rt3}{\small$_{\mathcal{C}_3}^{\mathcal{C}}\mathbf{T}$}
    \includegraphics[width=0.4\textwidth]{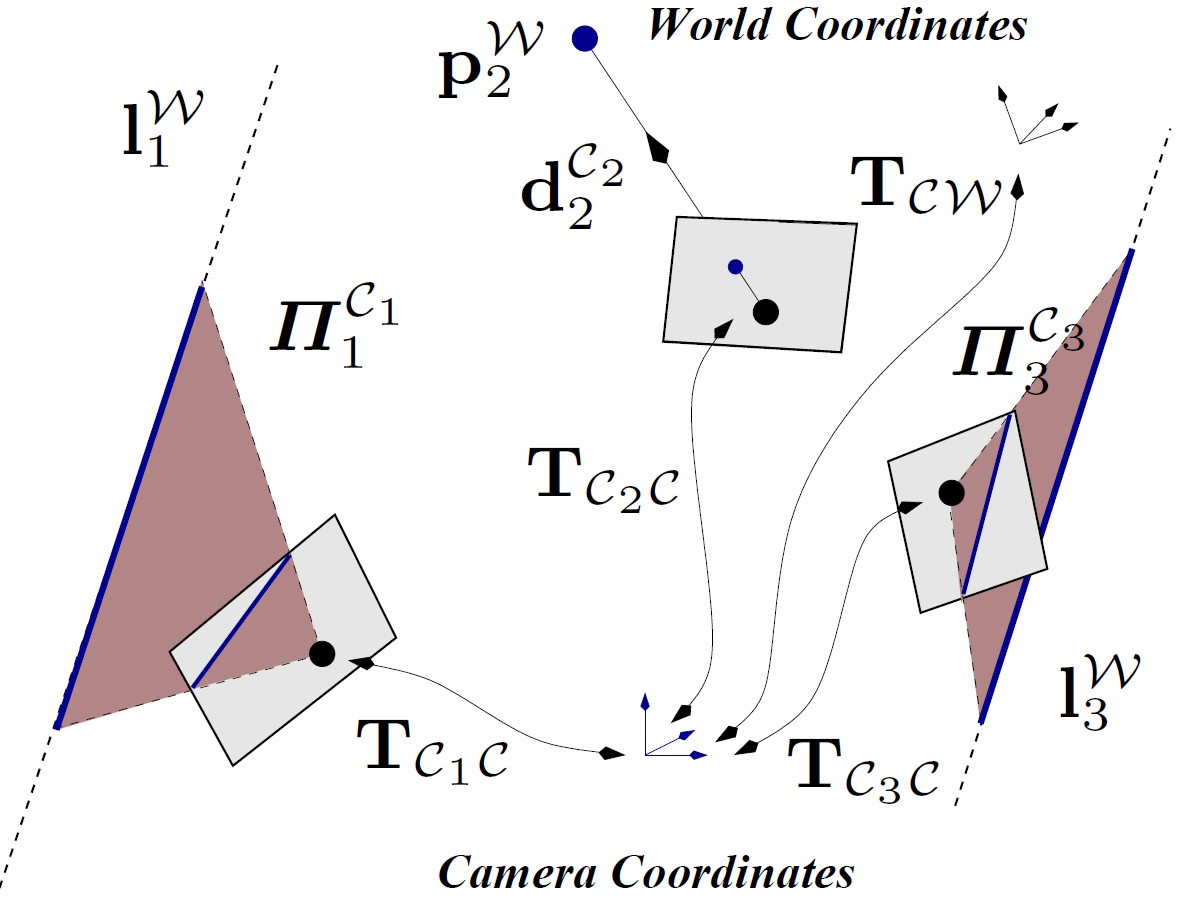}
    \label{fig:intro_1p2l}
    }
    \caption{Illustration of the two minimal problems solved in this paper. We estimate the transformation parameters $\mathbf{T}_{\mathcal{C}\mathcal{W}}$ in two different settings: the case where we have two 3D points, one 3D line, and their respective images~\protect\subref{fig:intro_2p1l}; and the case where we have two 3D lines, one 3D point and their respective images~\protect\subref{fig:intro_1p2l}.}
    \label{fig:intro}
\end{figure}

{\it 1) Camera Pose using 3D Points:~}
For a set of 3D points $\mathbf{p}_j^{\mathcal{W}}$ and their respective images, since the camera parameters are known, the pose for a multi-perspective camera is given by $\mathbf{T}_{\mathcal{C}\mathcal{W}}$, for a $\{ \mathbf{p}_j^{\mathcal{W}} \mapsto \mathbf{d}_j^{\mathcal{C}_i} \}$ for $j=1,\dots,N$, where $\mathbf{d}_j^{\mathcal{C}_i} \in \mathbb{R}^3$ is the inverse projection direction, given by the image of $\mathbf{p}_j^{\mathcal{W}}$ seen in the camera $\mathcal{C}_i$~\cite{hartley04,ma03}. Formally, the pose is given by the $\mathbf{T}_{\mathcal{C}\mathcal{W}}$, such that
\begin{equation}
    \label{eq:point_constraint}
    \mathbf{T}_{\mathcal{C}\mathcal{W}}
    \begin{bmatrix}
    \delta_j\  \mathbf{R}_{\mathcal{C}_i\mathcal{C}}\ \mathbf{d}_j^{\mathcal{C}_i} + \mathbf{c}_i^{\mathcal{C}} \\
    1
    \end{bmatrix}
    =
    \begin{bmatrix}
    \mathbf{p}_j^{\mathcal{W}}\\
    1
    \end{bmatrix}
    \ \ \text{for all } j=1,\dots,N ,
\end{equation}
where $\delta_j$ is an unknown depth of $\mathbf{p}_j^{\mathcal{C}_i}$, w.r.t. the camera center $\mathbf{c}_i^{\mathcal{C}}\in\mathbb{R}^3$.

{\it 2) Camera Pose using 3D Lines:~}
To represent 3D straight lines in the world we use {\it Pl\"ucker} coordinates \cite{pottmann01}, i.e. $\mathbf{l}_j^{\mathcal{W}} \dot{\sim}(\bar{\mathbf{l}}_j^{\mathcal{W}},\tilde{\mathbf{l}}_j^{\mathcal{W}})$ where $\bar{\mathbf{l}}_j^{\mathcal{W}},\tilde{\mathbf{l}}_j^{\mathcal{W}}\in\mathbb{R}^3$ are the line's direction and moment, respectively. Since the camera parameters are known, their respective images can be represented by an interpretation plane $\bm{\Pi}_j^{\mathcal{C}_i}\in\mathbb{R}^4 = (\bar{\bm{\pi}}_j^{\mathcal{C}_i}, \check{\pi}_j^{\mathcal{C}_i})$ \cite{hartley04,ma03}, where $\bar{\bm{\pi}}_j^{\mathcal{C}_i}$ is the normal vector to the plane and $\check{\pi}_j^{\mathcal{C}_i}$ is its distance to the origin of the respective coordinate system, which in this case is equal to zero (the interpretation plane passes through the center of the camera $\mathcal{C}_i$). Under the correct pose the 3D line lies on the interpretation plane formed by the corresponding 2D line and the camera center. Thus the required pose using lines is given by  $\mathbf{T}_{\mathcal{C}\mathcal{W}}$ for a $\{ \mathbf{l}_j^{\mathcal{W}} \mapsto \bm{\Pi}_j^{\mathcal{C}_i} \}$, such that
\begin{equation}
\label{eq:line_proj_constraint}
\underbrace{\begin{bmatrix}
\hat{\tilde{\mathbf{l}}}_j^{\mathcal{W}} & \bar{\mathbf{l}}_j^{\mathcal{W}} \\
\left.\bar{\mathbf{l}}_j^{\mathcal{W}}\right.^T & 0
\end{bmatrix}}_{\mathbf{L}_j^{\mathcal{W}}\in\mathbb{R}^{4\times 4}}\ 
\mathbf{T}^{-T}_{\mathcal{C}\mathcal{W}}\ \mathbf{T}^{-T}_{\mathcal{C}_i\mathcal{C}}\ \bm{\Pi}_j^{\mathcal{C}_i} = \mathbf{0},\ \ \text{for all } j=1,\dots,N ,
\end{equation}
where: $\mathbf{L}_j^{\mathcal{W}}$ is the {\it Pl\"{u}cker} matrix of the line $\mathbf{l}_j^{\mathcal{W}}$ \cite{pottmann01}; the hat represents skew-symmetric matrix that linearizes the external product, such that $\mathbf{a}\times \mathbf{b} = \hat{\mathbf{a}}\mathbf{b}$; and $N$ is the number of correspondences between 3D lines in the world and their respective interpretation planes. 


\subsection{Minimal Pose using Points and Lines}
\label{sec:minimal_absolute_pose}

Similar to the cases of using only points or lines, the minimal pose is computed by having three of these features in the world, and their respective images. This means that, for the minimal pose addressed in this paper, and according to Sec.~\ref{sec:intro}, there are two cases:
\begin{itemize}
    \item The estimation of $\mathbf{T}_{\mathcal{C}\mathcal{W}}$, knowing: $\mathbf{l}_1^{\mathcal{W}} \mapsto \bm{\Pi}_1^{\mathcal{C}_1}$; $\mathbf{p}_{2}^{\mathcal{W}} \mapsto \mathbf{d}_{2}^{\mathcal{C}_2}$; $\mathbf{p}_{3}^{\mathcal{W}} \mapsto \mathbf{d}_{3}^{\mathcal{C}_3}$; and $\mathbf{T}_{\mathcal{C}_i\mathcal{C}}$ for $i=1,2,3$. A graphical representation of this problem is shown in Fig.~\ref{fig:intro}\subref{fig:intro_2p1l}; and
    \item The estimation of $\mathbf{T}_{\mathcal{C}\mathcal{W}}$, knowing: $\mathbf{l}_1^{\mathcal{W}} \mapsto \bm{\Pi}_1^{\mathcal{C}_1}$; $\mathbf{p}_{2}^{\mathcal{W}} \mapsto \mathbf{d}_{2}^{\mathcal{C}_2}$; and $\mathbf{l}_3^{\mathcal{W}} \mapsto \bm{\Pi}_3^{\mathcal{C}_3}$; and $\mathbf{T}_{\mathcal{C}_i\mathcal{C}}$ for $i=1,2,3$. This problem is depicted in Fig.~\ref{fig:intro}\subref{fig:intro_1p2l}.
\end{itemize}

We show the solutions to these minimal problems in the next section.
\section{Solution to the Minimal Pose Problem using Points and Lines}
\label{sec:our_contrib}

As a first step, we transform the world and camera coordinate systems using predefined transformations to the data (Sec.~\ref{sec:predefined_transformations}). Note that we can first compute the pose in this new coordinate systems, and then recover the real pose in the original coordinate frames by using the inverse of the predefined transformations. The use of such predefined transformations can greatly simplify the underlying polynomial equations and enable us to develop low-degree polynomial solutions.

\subsection{Select the World and Camera Coordinate Systems}
\label{sec:predefined_transformations}
\begin{figure}[t]
\vspace{-0.3cm}
  \centering
  \subfloat[World coordinate system]{
    \psfrag{p2}{\small$\mathbf{p}_2^{\mathcal{W}}$}
    \psfrag{l1}{\small$\mathbf{l}_1^{\mathcal{W}}$}
    \psfrag{p2ez}{\small$\mathbf{p}_2^{\mathcal{W}} \equiv \bm{e}_z$}
    \psfrag{x}{\small$\bm{e}_x$}
    \psfrag{y}{\small$\bm{e}_y$}
    \psfrag{z}{\small$\bm{e}_z$}
    \includegraphics[height=0.14\textheight]{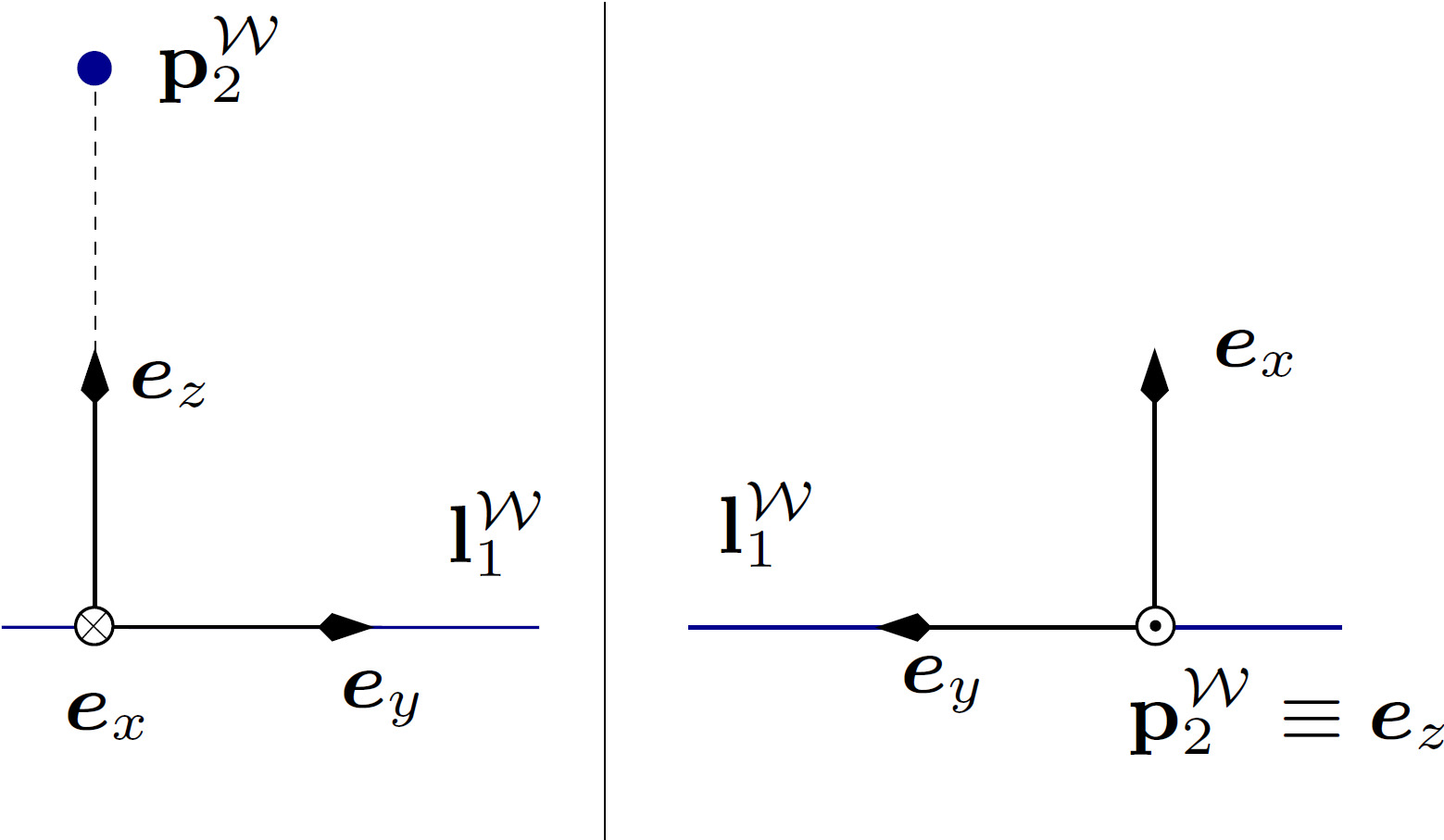}
    \label{fig:world_refs}
  }\qquad
  \subfloat[Camera coordinate system]{
    \psfrag{a}{\small$\gamma_z$}
    \psfrag{x}{\small$\bm{e}_x$}
    \psfrag{y}{\small$\bm{e}_y$}
    \psfrag{z}{\small$\bm{e}_z$}
    \psfrag{Pi}{\small$\bm{\Pi}_1^{\mathcal{W}}$}
    \includegraphics[height=0.14\textheight]{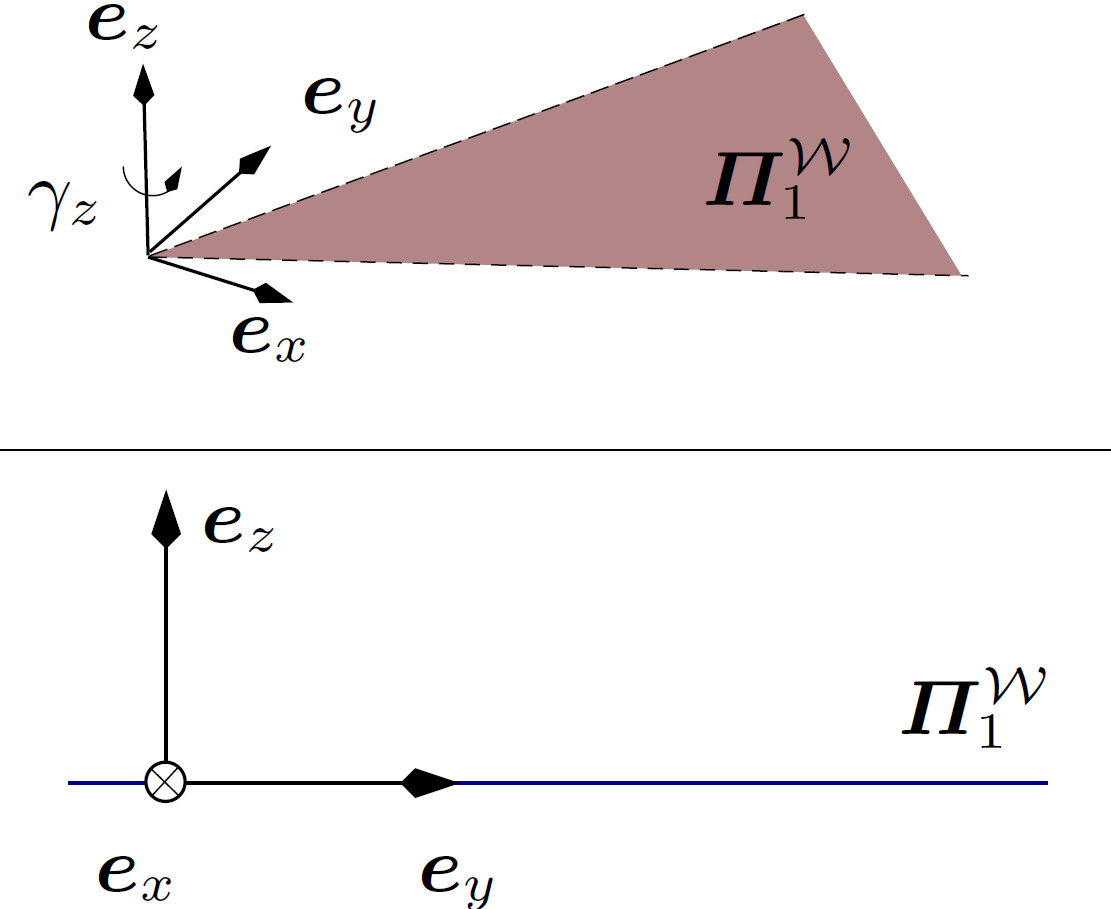}
    \label{fig:camera_refs}
  }
    \caption{Depiction of the selected world and camera coordinate systems. \protect\subref{fig:world_refs} shows the considered world coordinate system, while \protect\subref{fig:camera_refs} presents the selected camera coordinate system. We note that while the world coordinate system is uniquely defined, the camera coordinate system can be defined up to a $z$--axis rotation.}
    \label{fig:selec_refs}
\end{figure}

Let us consider initial transformations such that the data in the world coordinate system verify the following specifications:
\begin{itemize}
    \item Centered on the $\mathbf{l}_1^{\mathcal{W}}$;
    \item With the $y$--axis aligned with the line's direction; and
    \item Such that $\mathbf{p}_2^{\mathcal{W}} = \begin{bmatrix}0 & 0 & * \end{bmatrix}^T$, where $*$ takes any value in $\mathbb{R}$ .
\end{itemize}
A graphical representation of these specifications is shown in Fig.~\ref{fig:selec_refs}\subref{fig:world_refs}. Regarding the camera coordinate system, we aim at having the following specifications:
\begin{itemize}
    \item Centered in the $\mathcal{C}_1$; and
    \item With the $z$--axis aligned with the interpretation plane normal.
\end{itemize}
A graphical representation of the camera's coordinate system is shown in Fig.~\ref{fig:selec_refs}\subref{fig:camera_refs}. The predefined transformation can be computed easily and the details are shown in the supplementary material. In the following subsections we present our solutions to the minimal case. 

\subsection{Solution using two 3D points and a 3D straight line}
\label{sec:twopoints_oneline}

Here, we present a closed-form solution using 2 points and 1 line. The minimal pose is computed using the coplanarity constraint on the 3D line and its associated interpretation plane \eqref{eq:line_proj_constraint}, and using collinearity constraint associated with the point correspondences \eqref{eq:point_constraint}.

From the selected coordinate systems (Sec.~\ref{sec:predefined_transformations}), the rotation between the camera and world coordinate systems is given by
\begin{equation}
    \mathbf{R}_{\mathcal{C}\mathcal{W}} =
    \begin{bmatrix}
    c\theta & 0 & -s\theta \\
    0  & 1 &   0 \\
    s\theta & 0 &  c\theta
    \end{bmatrix}
    \begin{bmatrix}
    c\alpha  & s\alpha & 0 \\
    -s\alpha & c\alpha & 0 \\
      0 &  0 & 1
    \end{bmatrix},
\end{equation}
with unknowns $c\theta$ \& $s\theta$\footnote{To simplify, we denote $\cos(\theta)$ as $c\theta$ and $\sin(\theta)$ as $s\theta$, respectively.} and $c\alpha$ \& $s\alpha$. One can notice that, as a result of the predefined transformations, we have reduced one degree of freedom on the rotation matrix. In addition, one has
\begin{equation}
    \label{eq:constraint_translation}
    \mathbf{R}_{\mathcal{C}\mathcal{W}} \mathbf{t}_{\mathcal{C}\mathcal{W}} =
    \begin{bmatrix}
    \mathbf{*} \\ \mathbf{*} \\ 0
    \end{bmatrix},
    \text{where}\ *\ \text{can take any value in}\ \mathbb{R}.
\end{equation}
Now, let us consider the effect of collinearity associated with the first point correspondence. From~\eqref{eq:point_constraint}, we have
\begin{equation}
    \label{eq:effect_first_point}
    \mathbf{T}_{\mathcal{C}\mathcal{W}}\left(\delta_2\mathbf{d}_2^{\mathcal{C}} + \mathbf{c}_2^{\mathcal{C}}\right) = \mathbf{p}_2^{\mathcal{W}},
\end{equation}
where $\mathbf{d}_2^{\mathcal{C}} = \mathbf{R}_{\mathcal{C}_2\mathcal{C}}\mathbf{d}_2^{\mathcal{C}_2}$. Then, this can be solved as a function of the unknowns $t_1$, $t_2$, $t_3$, resulting in
\begin{align}
    \label{eq:2p1l_t1}
    t_1 & = \kappa_{1}^{3}[c\theta, s\theta, c\alpha, s\alpha, \delta_2];\\
    \label{eq:2p1l_t2}
    t_2 & = \kappa_{2}^{2}[c\alpha, s\alpha, \delta_2]; \text{and}\\
    \label{eq:2p1l_t3}
    t_3 & = \kappa_{3}^{3}[c\theta, s\theta, c\alpha, s\alpha, \delta_2],
\end{align}
where $\kappa_i^j[.]$ denotes the $i$\textsuperscript{th} polynomial equation, with degree $j$. The analytic representation of all coefficients is sent in the supplementary material.

Next, we take into account the effects of the second point:
\begin{equation}
    \label{eq:2p1l_second_point}
    \mathbf{T}_{\mathcal{C}\mathcal{W}}\left(\delta_3\mathbf{d}_3^{\mathcal{C}} + \mathbf{c}_3^{\mathcal{C}}\right) = \mathbf{p}_3^{\mathcal{W}}.
\end{equation}
Replacing the unknowns $t_1$, $t_2$, and $t_3$ in \eqref{eq:2p1l_second_point} by the results of \eqref{eq:2p1l_t1}-\eqref{eq:2p1l_t3}, we get three constraints on the unknowns $\theta$, $\alpha$, $\delta_2$, and $\delta_3$, such that
\begin{align}
    \label{eq:2p1l_second_point_constraints_1}
    \kappa_{4}^{3}[c\theta, s\theta, c\alpha, s\alpha, \delta_2, \delta_3] & = 0;\\
    \label{eq:2p1l_second_point_constraints_2}
    \kappa_{5}^{2}[c\alpha, s\alpha, \delta_2, \delta_3] & = 0;\ \text{and}\\
    \label{eq:2p1l_second_point_constraints_3}
    \kappa_{6}^{3}[c\theta, s\theta, c\alpha, s\alpha, \delta_2, \delta_3] & = 0.
\end{align}
In addition, considering the third row of the constraint defined in \eqref{eq:constraint_translation} and replacing $t_1$, $t_2$, and $t_3$ in this equation by the results of \eqref{eq:2p1l_t1}-\eqref{eq:2p1l_t3}, we obtain the following constraint
\begin{equation}
    \label{eq:2p1l_translation_constraints}
    \kappa_{7}^{3}[c\theta, s\theta, \delta_2] = 0.
\end{equation}

Now, solving \eqref{eq:2p1l_second_point_constraints_1}-\eqref{eq:2p1l_second_point_constraints_3}, and \eqref{eq:2p1l_translation_constraints} as a function of $c\theta$, $s\theta$, $c\alpha$, and $s\alpha$, we get
\begin{equation}
    \label{eq:2p1l_solutions_angles}
    c\theta = \frac{\kappa_{8}^{1}[\delta_2]}{\kappa_{9}^{2}[\delta_2,\delta_3]},
    \ s\theta = \frac{\kappa_{10}^{1}[\delta_2,\delta_3]}{\kappa_{11}^{2}[\delta_2,\delta_3]}, 
    \ c\alpha = \frac{\kappa_{12}^{2}[\delta_2,\delta_3]}{\kappa_{13}^{2}[\delta_2,\delta_3]},
    \ \text{and}
    \ s\alpha = \frac{\kappa_{14}^{2}[\delta_2,\delta_3]}{\kappa_{15}^{2}[\delta_2,\delta_3]}.
\end{equation}
which we replace in the trigonometric relations $c\theta^2 + s\theta^2 -1 = 0$ and $c\alpha^2 + s\alpha^2 -1 = 0$, getting two constraints of the form
\begin{equation}
    \frac{\kappa_{16}^{2}[\delta_2,\delta_3]}{\kappa_{17}^{2}[\delta_2,\delta_3]} = 0 \ \ \text{and} \ \
    \frac{\kappa_{18}^{2}[\delta_2,\delta_3]}{\kappa_{19}^{2}[\delta_2,\delta_3]} = 0.
\end{equation}
However, solving the above equations as a function of the unknowns $\delta_2$ and $\delta_3$ is the same as solving
\begin{equation}
    \label{eq:solution_pose_2p1l}
    \kappa_{16}^{2}[\delta_2,\delta_3] = \kappa_{18}^{2}[\delta_2,\delta_3] = 0,
\end{equation}
that corresponds to the estimation of the intersection points between two quadratic curves which, according to {\it B\'{e}zout's} theorem \cite{cox15}, has four solutions. There are many generic solvers in the literature to compute these solutions (such as \cite{stewenius05,kukelova08,kukelova12,larsson16,larsson17}). However, since we are dealing with very simple polynomial equations, we derive our own fourth degree polynomial. From \eqref{eq:solution_pose_2p1l}, solving one polynomial as a function of $\delta_2$ and replacing these results in the other (the square root is removed using simple algebraic manipulations), we get
\begin{align}
    \label{eq:pose_2p1l_computation_1}
    \kappa_{19}^{4}[\delta_2] = 0 \ \ \text{and} \ \ &  \\
    \label{eq:pose_2p1l_computation_2}
    \delta_3 = \frac{\kappa_{20}^{1}[\delta_2] \pm \sqrt{\kappa_{21}^{2}[\delta_2]}}{\kappa_{22}^{1}[\delta_2]} &.
\end{align}
Details on these derivations are provided in the suplementary material. Finally, to compute the pose, one has to solve \eqref{eq:pose_2p1l_computation_1}, which can be computed in a closed-form (using Ferrari's formula), getting up to four real solutions for $\delta_2$. Then, by back-substituting $\delta_2$ in \eqref{eq:pose_2p1l_computation_2} we get the respective solutions for $\delta_3$ (notice that from the two possible solutions for $\delta_3$ one will be trivially ignored, since \eqref{eq:solution_pose_2p1l} can have only up to four solutions). The pair $\{ \delta_2, \delta_3 \}$ is afterwards used in \eqref{eq:2p1l_solutions_angles} to compute the respective $\{ c\theta, s\theta, c\alpha, s\alpha \}$, and then in \eqref{eq:2p1l_t1}-\eqref{eq:2p1l_t3} to estimate $\{t_1, t_2, t_3\}$.

\subsection{Solution using two 3D straight lines and a 3D point}
\label{sec:twolines_onepoint}

This subsection presents the solution to the multi-perspective pose problem using 2 lines and 1 point.
As before, we consider the predefined transformations to the input data defined in Sec.~\ref{sec:predefined_transformations}, which already includes the coplanarity constraint associated with the first 3D line. Under these assumptions, we start by considering the collinearity constraint associated with the 3D point and its respective image~\eqref{eq:point_constraint} and, then, use the coplanarity constraint of the second 3D line and its respective interpretation plane \eqref{eq:line_proj_constraint}.

We start by using the same steps of Sec.~\ref{sec:twopoints_oneline}, i.e. we get the translation parameters as a function of $c\theta$, $s\theta$, $c\alpha$, $s\alpha$, and $\delta_2$, which are given by \eqref{eq:2p1l_t1}-\eqref{eq:2p1l_t3}. Then, we replace the translation parameters in the third row of \eqref{eq:constraint_translation} by the results of \eqref{eq:2p1l_t1}-\eqref{eq:2p1l_t3}, which gives \eqref{eq:2p1l_translation_constraints}. Afterwards, we solve \eqref{eq:2p1l_translation_constraints} and the trigonometric constraint $c\theta^2 + s\theta^2 - 1 = 0$, as a function of $c\theta$ and $s\theta$, resulting in
\begin{equation}
    \label{eq:pose_1p2l_cstheta}
    c\theta = \kappa_{23}^{1}[\delta_2] \ \ \ \text{and} \ \ \
    s\theta = \pm\sqrt{\kappa_{24}^{2}[\delta_2]}.
\end{equation}

Now, we consider the constraints associated with the second line which, since $\mathbf{T}_{\mathcal{C}_3\mathcal{C}}$ is known, is given by
\begin{equation}
\label{eq:pose_1p2l_2nd_line}
\mathbf{L}_3^{\mathcal{W}}\ 
\mathbf{T}^{-T}_{\mathcal{C}\mathcal{W}}\ \bm{\Pi}_3^{\mathcal{C}} = \mathbf{0},
\end{equation}
where $\bm{\Pi}_3^{\mathcal{C}} = \mathbf{T}^{-T}_{\mathcal{C}_3\mathcal{C}}\ \bm{\Pi}_3^{\mathcal{C}_3}$. Replacing the translation parameters in the above equations by the results of \eqref{eq:2p1l_t1}-\eqref{eq:2p1l_t3}, and $c\theta$ by the outcome of \eqref{eq:pose_1p2l_cstheta} (notice that, for now we keep the unknown $s\theta$), we get four polynomial equations with degree two, as a function of variables $\delta_2$, $s\alpha$, $c\alpha$, and $s\theta$. Solving two of them as a function of $c\alpha$ and $s\alpha$, we get
\begin{equation}
\label{eq:pose_1p2l_csalpha}
c\alpha = \frac{\kappa_{25}^{2}[s\theta,\delta_2]}{\kappa_{26}^{1}[s\theta,\delta_2]}  \ \ \ \text{and} \ \ \ s\alpha = \frac{\kappa_{27}^{2}[s\theta,\delta_2]}{\kappa_{28}^{1}[s\theta,\delta_2]}.
\end{equation}
Now, replacing these results into the trigonometric relation $c\alpha^2 + s\alpha^2 - 1 = 0$, we get a constraint of the form
\begin{equation}
    \frac{ \kappa_{29}^{4}[s\theta,\delta_2] }{ \kappa_{30}^{2}[s\theta,\delta_2] } = 0 \ \ \Rightarrow \ \ \kappa_{29}^{4}[s\theta,\delta_2] = 0.
\label{eq:salpha_delta2}
\end{equation}
Notice that, from \eqref{eq:pose_1p2l_cstheta}, the expression that defines $s\theta$, as a function of $\delta_2$, has a square root of a polynomial equation. Then, starting from \eqref{eq:salpha_delta2}, we simplify the problem by: 1) taking the terms with $s\theta$ to the right side of the equation:
\begin{equation}
    \kappa_{29}^{4}[s\theta,\delta_2] = 0 \Rightarrow s\theta^4 + \kappa_{31}^{2}[\delta_2] s\theta^2 + \kappa_{32}^{4}[\delta_2] = - \left(\kappa_{33}^{1}[\delta_2] s\theta^2 + \kappa_{34}^{3}[\delta_2] \right) s\theta ;
\end{equation}
2) squaring both sides \& moving all the terms to the left side of the equation:
\begin{equation}
    s\theta^8 + \kappa_{35}^{2}[\delta_2] s\theta^6 + \kappa_{36}^{6}[\delta_2] s\theta^4 + \kappa_{37}^{6}[\delta_2] s\theta^2 + \kappa_{38}^{8}[\delta_2] = 0 ;
\end{equation}
and, finally, 3) replacing $s\theta$ using \eqref{eq:pose_1p2l_cstheta} (notice that the square root and the $\pm$ signal is removed), we get
\begin{equation}
    \label{eq:solve_1p2l}
    \kappa_{39}^{8}[\delta_2] = 0,
\end{equation}
which as up to eight real solutions. To get the pose: 1) we compute $\delta_2$ from the real roots of \eqref{eq:solve_1p2l}; 2) for each $\delta_2$, we get $\{ c\theta, s\theta\}$ from \eqref{eq:pose_1p2l_cstheta}; 3) we compute $\{ c\alpha, s\alpha\}$ from \eqref{eq:pose_1p2l_csalpha}; and 4) by back-substituting all these unknowns, we get $\{ t_1, t_2, t_3\}$ from \eqref{eq:2p1l_t1}-\eqref{eq:2p1l_t3}, obtaining the estimation of the camera pose.
\section{Experimental Results} \label{sec:exp}
In these experiments, we consider the methods proposed in Sec.~\ref{sec:our_contrib} and existing multi-perspective algorithms to solve the pose using three points \cite{gim15} or three lines \cite{gim16}. All algorithms were implemented in {\tt MATLAB} and are available in the author's website.

We start by using synthetic data (Sec.~\ref{sec:exp_syn}): 1) we evaluate the number of real solutions and analyze their computational complexity; and 2) we test each method with noise.
Next, we show results using real data: 1) we evaluate the minimal solutions in a RANSAC framework (Sec.~\ref{sec:RANSAC}); and 2) we use each method in a 3D path reconstruction using a real multi-perspective camera (Sec.~\ref{sec:real}).

\subsection{Results with Synthetic Data}
\label{sec:exp_syn}

To get the data, we randomly define the ground truth camera pose, $\mathbf{T}_{GT}$. Three perspective cameras were generated (randomly distributed in the environment) in which their position w.r.t the camera coordinate system is assumed to be known, $\mathbf{T}_{\mathcal{C}_i\mathcal{C}}$. Then, for each camera ${\mathcal{C}_i}$, we define a feature in the world, and their projection into the image:
\begin{description}
    \item[$\mathbf{p}_i^{\mathcal{W}} \mapsto \mathbf{d}_i^{\mathcal{C}_i}$:] Points in the world $\mathbf{p}_{i}^{\mathcal{W}}$ are projected into the image $\mathbf{u}_i^{\mathcal{I}_i}$, by using a predefined calibration matrix. We had noise in the image pixels, and get the corresponding 3D inverse projection direction $\mathbf{d}_i^{\mathcal{C}_i}$.
    \item[$\mathbf{l}_i^{\mathcal{W}} \mapsto \bm{\Pi}_i^{\mathcal{C}_i}$:]  3D points defining the edges of the 3D line $\mathbf{l}_{i}^{\mathcal{W}}$ are projected into the image, $\{\mathbf{u}_{1,i}^{\mathcal{I}_i},\ \mathbf{u}_{2,i}^{\mathcal{I}_i}\}$. To each image point of the edge, we add noise (as we did in the previous point) and compute the respective inverse projection directions $\{\mathbf{d}_{1,i}^{\mathcal{C}_i}, \mathbf{d}_{2,i}^{\mathcal{C}_i}\}$. The interpretation plane is given by $\bm{\Pi}_i^{\mathcal{C}_i} = \begin{bmatrix} \mathbf{d}_{1,i}^{\mathcal{C}_i} \times \mathbf{d}_{2,i}^{\mathcal{C}_i}& 0\end{bmatrix}$.
\end{description}
After getting the data, we apply the known transformations $\mathbf{T}_{\mathcal{C}_i\mathcal{C}}$ to obtain the corresponding features in the global camera coordinate system, such that
\begin{description}
    \item[$\mathbf{p}_i^{\mathcal{W}} \mapsto \mathbf{c}_i^{\mathcal{C}} + \delta_i \mathbf{d}_i^{\mathcal{C}}$] in which $\mathbf{c}_i^{\mathcal{C}}$ is the perspective camera center and $\mathbf{d}_i^{\mathcal{C}} = \mathbf{R}_{\mathcal{C}_i\mathcal{C}}\mathbf{d}_i^{\mathcal{C}_i}$.
    \item[$\mathbf{l}_i^{\mathcal{W}} \mapsto \bm{\Pi}_i^{\mathcal{C}}$] in which $\bm{\Pi}_i^{\mathcal{C}} = \mathbf{T}^{-T}_{\mathcal{C}_i\mathcal{C}}\bm{\Pi}_i^{\mathcal{C}_i}$.
\end{description}

\begin{figure}[t]
    \vspace{-0.3cm}
  \centering
    \subfloat[Numerical Results.]{
    \includegraphics[height=0.105\textheight]{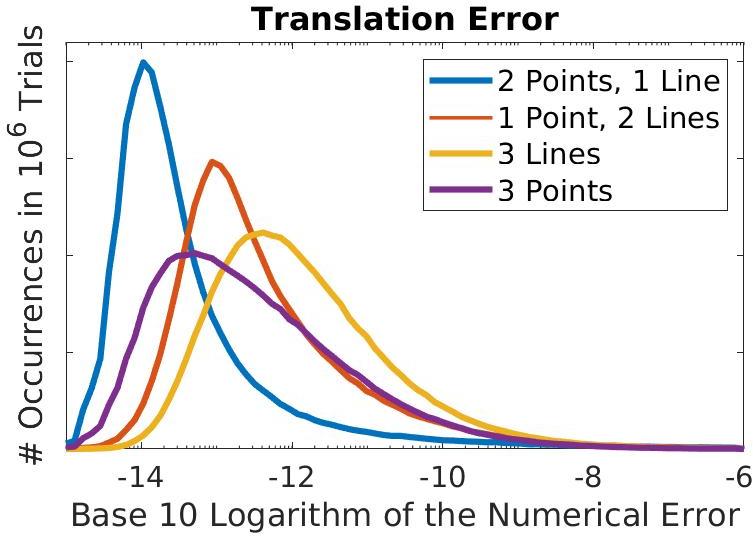}
    \includegraphics[height=0.105\textheight]{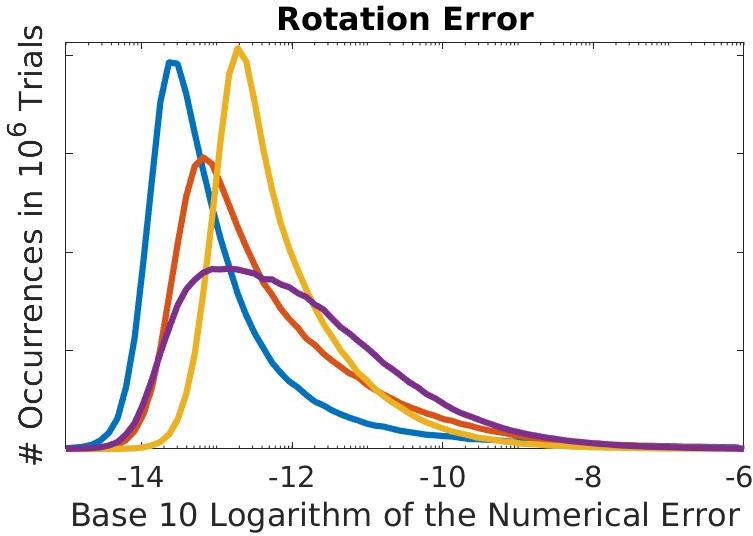}
    \label{fig:num_results:num}
    }
    \subfloat[Number of Occurrences.]{
    \includegraphics[height=0.105\textheight]{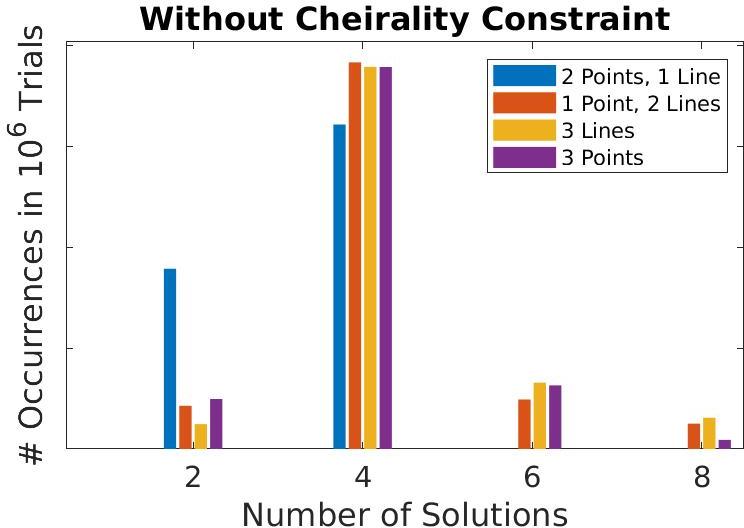}
    \includegraphics[height=0.105\textheight]{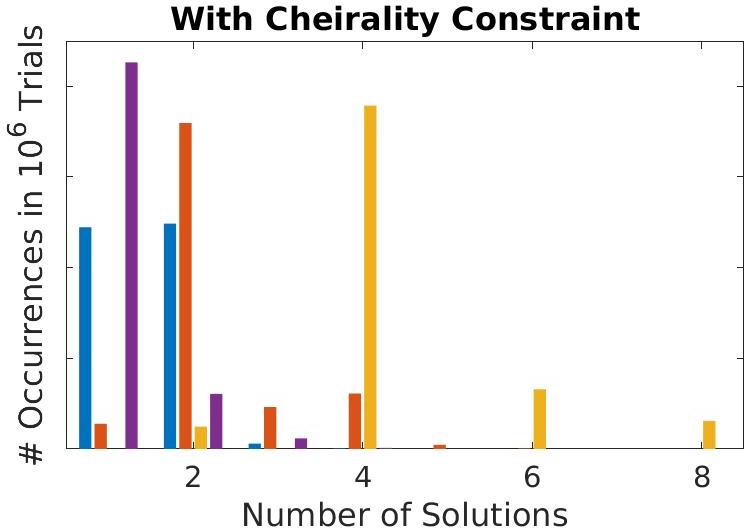}
    \label{fig:num_results:nocc}
    }\\ 
    \subfloat[The total and median of the computation time for solving the camera poses.]{
    \begin{tabular}{|r||c|c|c|c|}
    \hline
    {\bf Algorithm} & ~2 Points \& 1 Line~ & ~1 Point \& 2 Lines~ & ~3 Points \cite{gim15}~ & ~3 Lines \cite{gim16}~ \\ \hline \hline
    {\bf Total}  & $151.3 s$ & $516.6 s$ & $512.2 s$ & $1481.7 s$ \\ \hline
    {\bf Median} & $135 \mu s$& $439\mu s$ & $480\mu s$ & $1161\mu s$ \\ \hline
    \end{tabular}
    \label{fig:num_results:tc}
    }
    \caption{Results obtained with numerical errors for the pose estimation, using the methods proposed in this paper ({\tt 2 Points, 1 Line} and {\tt 1 Point, 2 Lines}) and existing solutions for points ({\tt 3 Points}) and lines ({\tt 3 Lines}).}
    \label{fig:num_results}
\end{figure}

For the evaluation, we start by running an experiment in which we consider $10^6$ random trials, without adding noise in the image pixels. We use both methods presented in this paper, as well as the algorithms presented in \cite{gim15,gim16}. For each trial/method, in which $\mathbf{R}_{\mathcal{C}\mathcal{W}}$ and $\mathbf{t}_{\mathcal{C}\mathcal{W}}$ are the estimated rotation and translation parameters, we:
1) compute the relative rotation that caused the deviation of the estimated rotation w.r.t. the ground-truth $\Delta \mathbf{R} = \mathbf{R}_{\mathcal{C}\mathcal{W}}\mathbf{R}_{GT}^T$, which can be represented by an axis-angle rotation, and the error is set as the respective angle in degrees; and
2) set $\|\mathbf{t}_{\mathcal{C}\mathcal{W}} - \mathbf{t}_{GT}\|$ as the translation error\footnote{For the cases in which the algorithms return multiple solutions, it was considered the cases with the smallest error using these metrics.}.  Fig.~\ref{fig:num_results}\subref{fig:num_results:num} shows that the methods are very similar in terms of numerical evaluation.

{\it Cheirality Constraint:~} All of the methods evaluated in this section produce multiple solutions for the pose. Our methods of Sec.~\ref{sec:twopoints_oneline} and Sec.~\ref{sec:twolines_onepoint} give up to four and eight solutions respectively. We discard imaginary solutions and the ones that are not physically realizable. The so-called {\it cheirality} constraint \cite{ma03} restricts points behind the camera (this is only possible to check in the cases in which we use point correspondences). We obtain the result for the $10^6$ trials with and without the {\it cheirality} constraint. Fig.~\ref{fig:num_results}\subref{fig:num_results:nocc} shows that the number of valid solutions (with the {\it cheirality} constraint) is lower for the algorithms that use more points.

{\it Computation Time:~} To conclude these tests, we present the evaluation of the computation time, required for each algorithm to compute all the $10^{6}$ trials. In theory, the method presented in Sec.~\ref{sec:twopoints_oneline} is the fastest, since it is computed in closed-form. On the other hand, both our method presented in Sec.~\ref{sec:twolines_onepoint} and the case of three points \cite{gim15} require the computation of the roots of an eighth degree polynomial equation, which requires iterative techniques. Moreover, the case of three lines \cite{gim16} not only requires the computation of an eighth degree polynomial equation, but also the computation of the null-space of a $3\times 9$ matrix, that also slows down the execution time. Results shown in Tab.~\ref{fig:num_results}\subref{fig:num_results:tc} validate the above assumptions. Note that these timing analysis are done using {\tt Matlab}, and porting the code to {\tt C++} would produce further speedup. 

\label{sec:num_errors}
\begin{figure}[t]
  \centering
    \includegraphics[height=0.105\textheight]{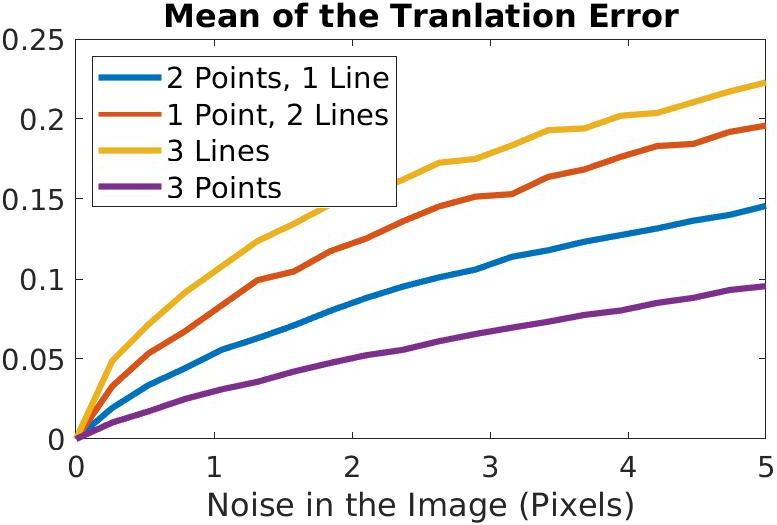} \hfill
    \includegraphics[height=0.105\textheight]{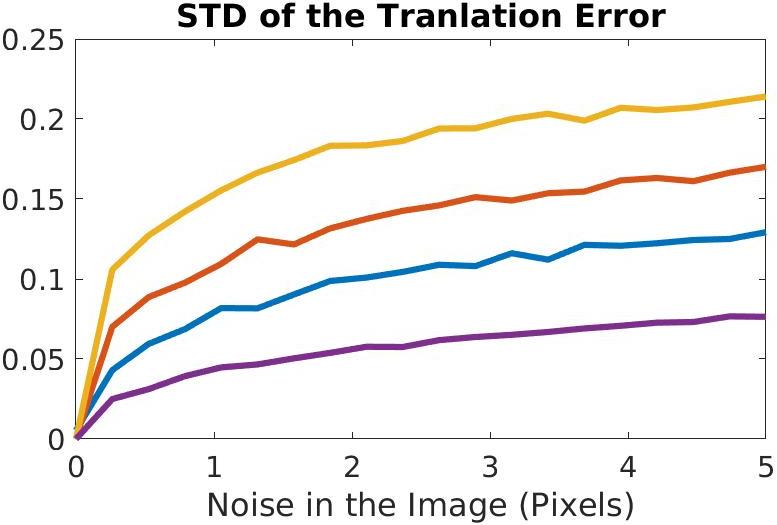} \hfill
    \includegraphics[height=0.105\textheight]{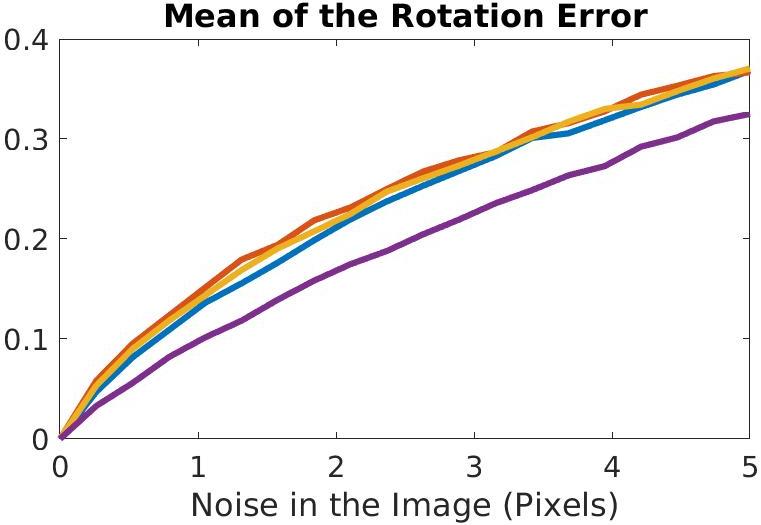}\hfill
    \includegraphics[height=0.105\textheight]{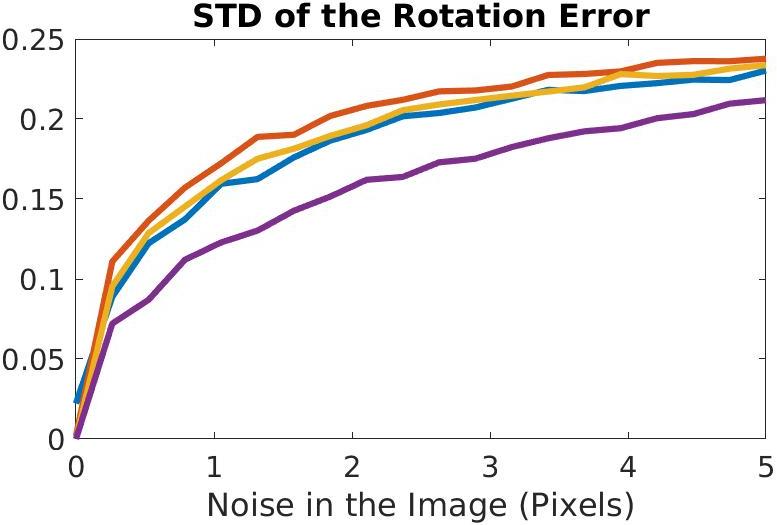}
    \caption{Comparative results for the methods using different type of features, as a function of the noise in the image pixels. {\tt 2 Points, 1 Line} and {\tt 1 Point, 2 Lines} show the results for the methods presented in this paper, while {\tt 3 Points} and lines {\tt 3 Lines} are techniques proposed in \cite{gim15,gim16}.}
    \label{fig:noise_results}
\end{figure}

Next, we evaluate the robustness of the methods in terms of image noise. For that purpose, we consider the same data-set generated, but we add noise in the image varying from 0 to 5 pixels. For each level of noise, we get $10^3$ random trials, compute the pose for all the four algorithms (notice the data required for each of the algorithms is different), and extract the average and standard deviation for all the $10^3$ trials for each level of noise. The results shown in Fig.~\ref{fig:noise_results} indicate that the algorithms that use more points are more robust to the noise.

\subsection{Evaluating Minimal Solutions in a RANSAC Framework}
\label{sec:RANSAC}
For these experiments, using real data, we evaluate the results of the minimal solutions in a RANSAC framework \cite{fischler81,nister03}. Since we are using points and lines correspondences, one needs to define two metrics for the re-projection errors: 1) For points, we use the geometric distance between the known pixels and the re-projection of 3D points using the estimated camera pose; and 2) For lines, we use the result presented in \cite{hofer17} which, for a ground truth line $\mathbf{l}_{GT}$ and a re-projected line $\mathbf{l}$ (both in the image), is given by
\begin{equation}
    d_L(\mathbf{l}_{GT},\mathbf{l})^{2} = \left( d_P(\mathbf{u}_1,\mathbf{l})^2 + d_P(\mathbf{u}_2,\mathbf{l})^2 \right)\text{exp}(2 \angle (\mathbf{l}_{GT},\mathbf{l})),
\end{equation}
where $d_P(.)$ denotes the geometric distance between a point and line in the image, and $\mathbf{u}_1$ \& $\mathbf{u}_2$ are the end points of $\mathbf{l}_{GT}$.

Then, we use a data-set from the ETH3D Benchmark~\cite{ETHz3D}. The data-set gives us the calibration and poses from a set of cameras, and the 3D points and their correspondences in the images. To extract the 3D lines and their correspondences in the images, we use the camera calibration \& pose parameters from the data-set and the {\tt Line3D++} algorithm~\cite{hofer17}.

\begin{figure}[t]
\vspace{-0.3cm}
  \centering
  \begin{tabular}[b]{c}%
    \subfloat[Examples of images and 2D Data used in this experimental results (blue lines and red points).]{
        \includegraphics[height=0.07\textheight]{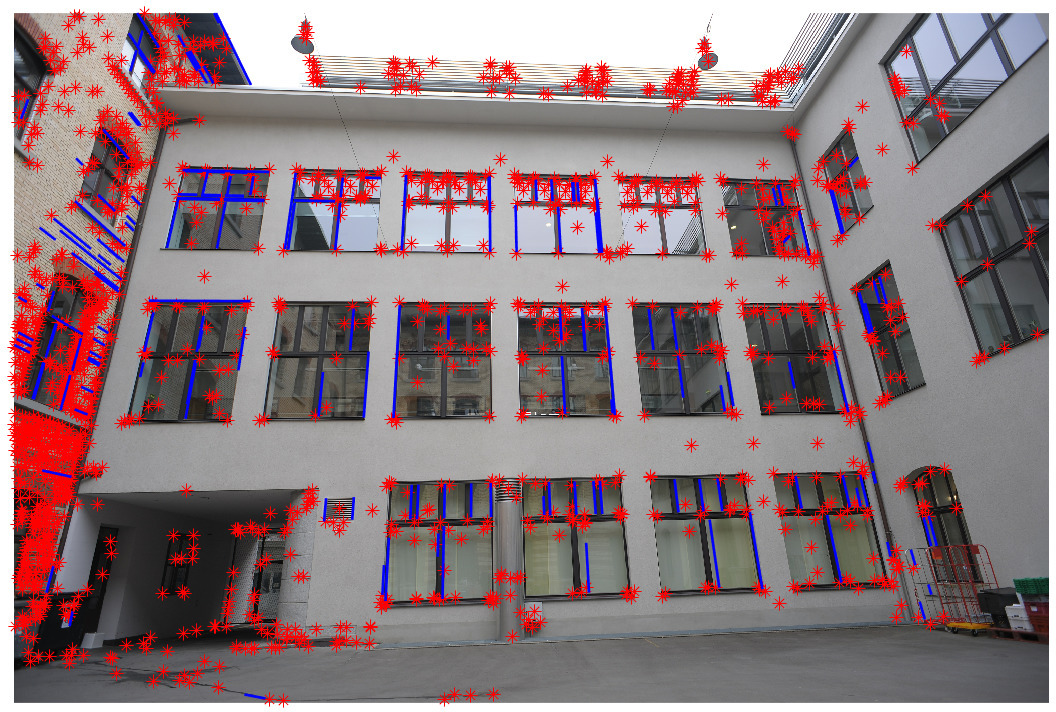}
        \includegraphics[height=0.07\textheight]{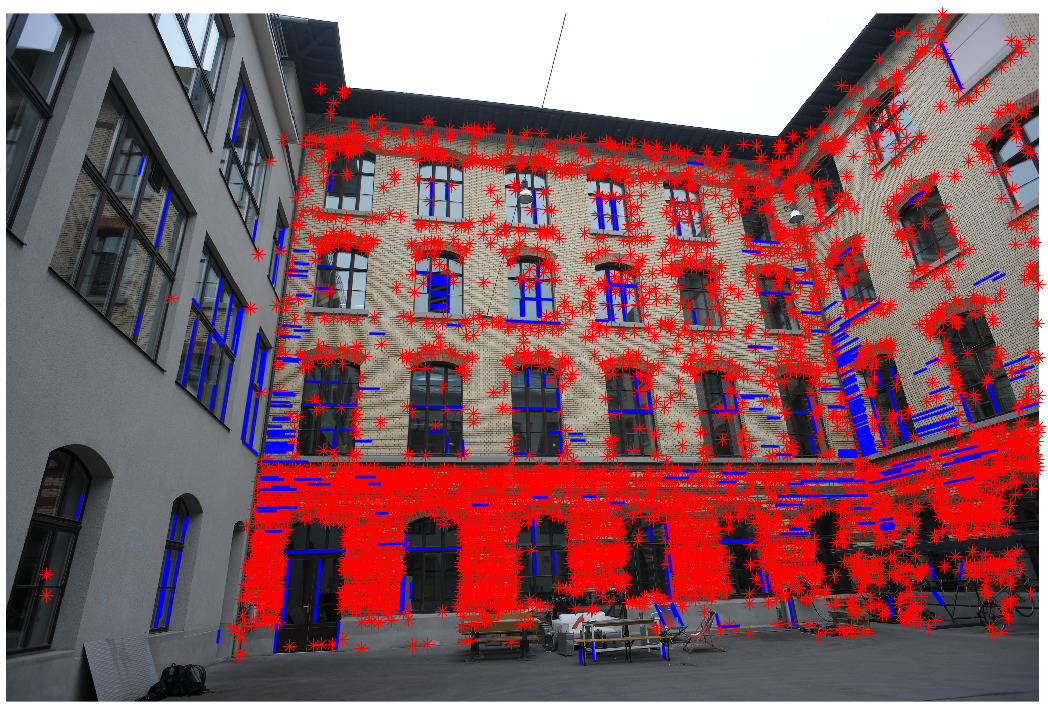}
        \includegraphics[height=0.07\textheight]{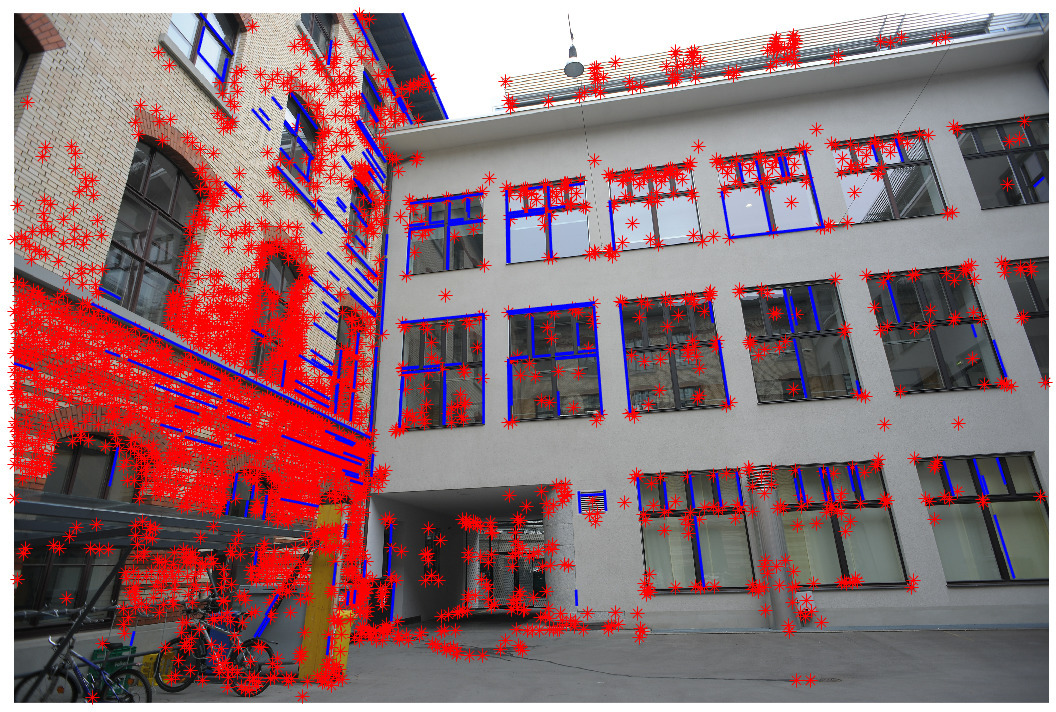}
        \label{fig:real_data_image}
    }\\
    \subfloat[3D Data set used in this experiments (blue lines and red points), and the estimated camera positions (green).]{
        \includegraphics[height=0.22\textheight]{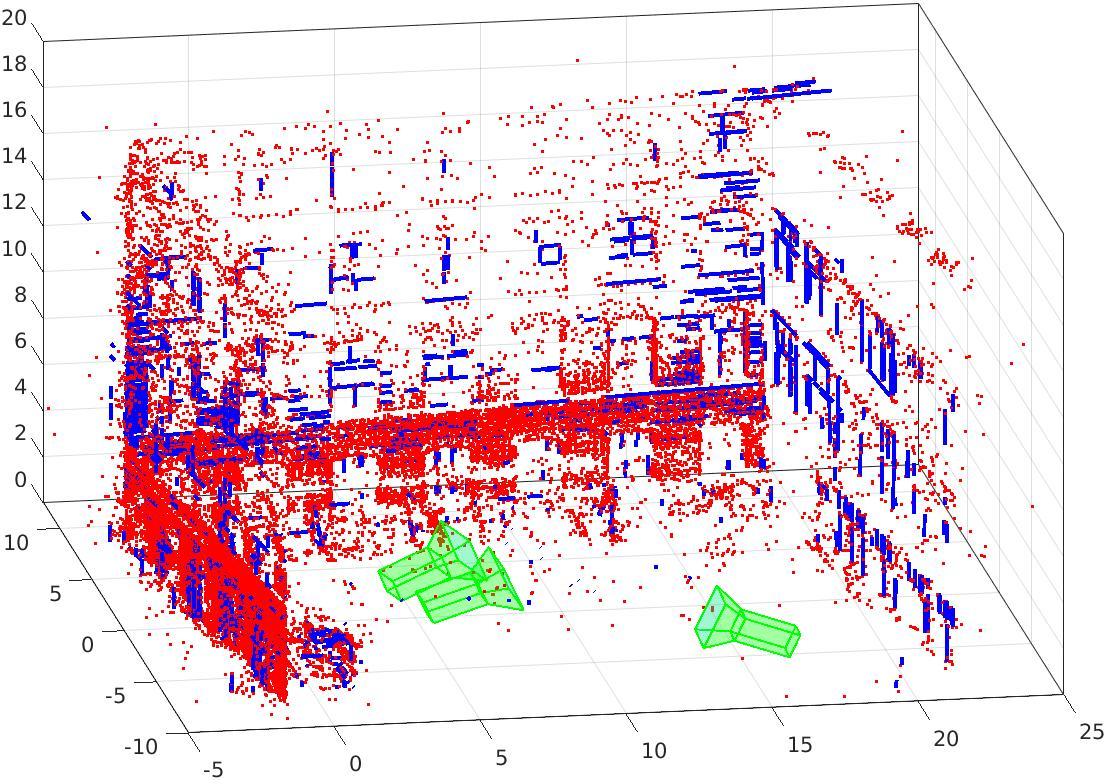}
        \label{fig:real_data_world}
    }
    \end{tabular}\hfill
    \begin{tabular}[b]{c}%
    \subfloat[Minimal solution in a RANSAC framework: varying the required number of inliers to stop the cycle.]{
        \includegraphics[height=0.140\textheight]{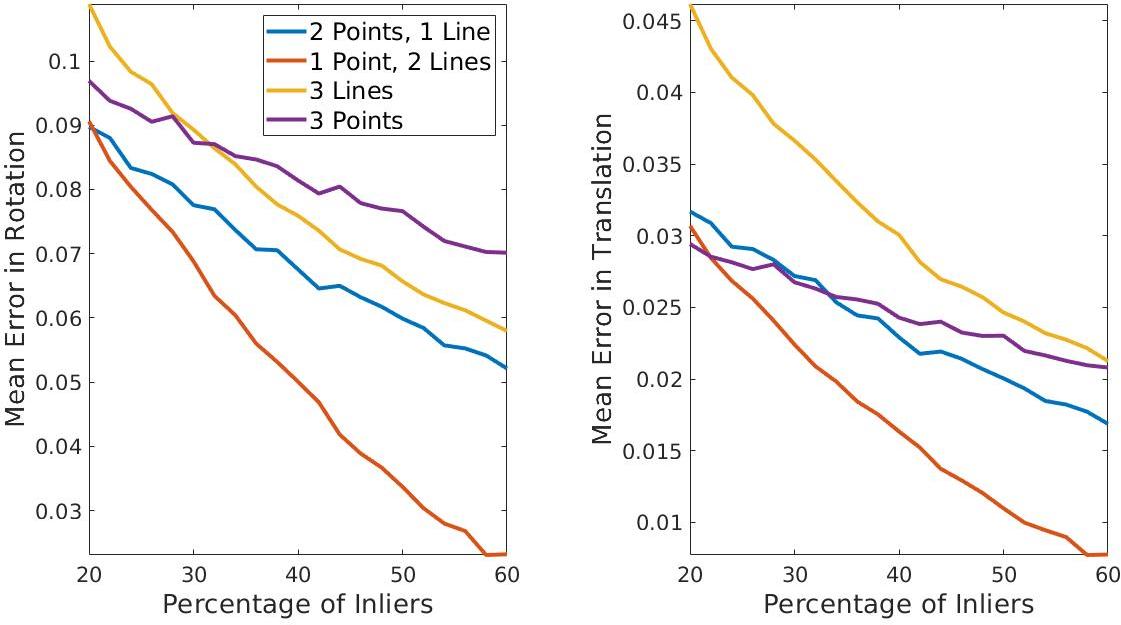}
        \label{fig:real_results_ninliers}
    }
    \\
    \subfloat[Minimal solution in a RANSAC framework: varying the thresholds to stop the cycle.]{
        \includegraphics[height=0.145\textheight]{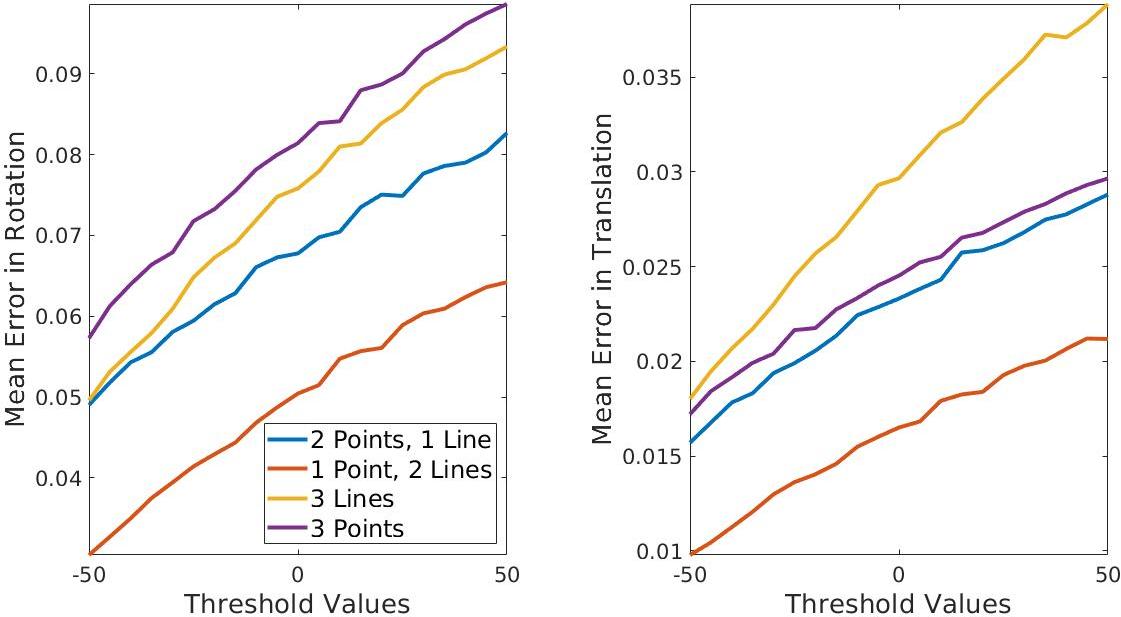}
        \label{fig:real_results_nthres}
    }
    \end{tabular}
    \caption{Evaluation of the proposed techniques and existing methods. As the evaluation criteria, we consider the required number of inliers and threshold to stop the RANSAC cycle. The errors in terms of rotation and translation parameters are afterwards computed, and compared between all the methods. \protect\subref{fig:real_data_image} and \protect\subref{fig:real_data_world} show three views and the 2D-3D data (points and lines) used in these experiments. \protect\subref{fig:real_results_ninliers} and \protect\subref{fig:real_results_nthres} show the proposed evaluation.}
    \label{fig:real}
\end{figure}

Examples of features in the image and its respective coordinates in the world are shown in Fig.~\ref{fig:real}\subref{fig:real_data_image} and~\ref{fig:real}\subref{fig:real_data_world}, respectively.
We run two experiments with these data, using both our methods and \cite{gim15,gim16}, under a RANSAC framework.
We start by defining a threshold for points and lines\footnote{Notice these thresholds must be different because of the differences between the metrics presented above.}.
To fairly select these thresholds, we run \cite{gim15,gim16} (that use solely points or lines respectively), and calibrate the values to ensure similar results in terms of errors as a function of the required number of inliers.
We use these line and point thresholds in our techniques.
Then, we vary the required number of inliers (a percentage of the all points and lines in the image), and for each we run the methods $10^4$ times.
In Fig.~\ref{fig:real}\subref{fig:real_results_ninliers} we show the results for the errors (using the metrics presented in Sec.~\ref{sec:exp_syn} for the translation and rotation errors), as a function of the percentage of inliers.

To conclude these experiments, we do some tests varying the threshold, for a fixed number of required inliers (in this case we consider 40 percent of the data), in a RANSAC framework.
To vary the threshold, we start from the values indicated in the previous paragraph, and vary as a function of the percentage of the corresponding values.
The results are shown in Fig.~\ref{fig:real}\subref{fig:real_results_nthres}, for thresholds ranging from 50 to 150 percent of the original threshold.

{\it Positives:~} As it can be seen from the results of Figs.~\ref{fig:real}\subref{fig:real_results_ninliers} and~\ref{fig:real}\subref{fig:real_results_nthres}, for the threshold values previously defined, both methods using three points and three lines have similar results, and, when comparing to the results of our solutions (using 2 points \& 1 line and 1 point \& 2 lines) one can see that the errors on the rotation and translation parameters are in general significantly lower.

\begin{figure}[t]
  \centering
  {
  \captionsetup[subfloat]{farskip=0pt,captionskip=0pt}
    \begin{tabular}[b]{c}%
    \subfloat[System.]{
    \includegraphics[height=0.194\textheight]{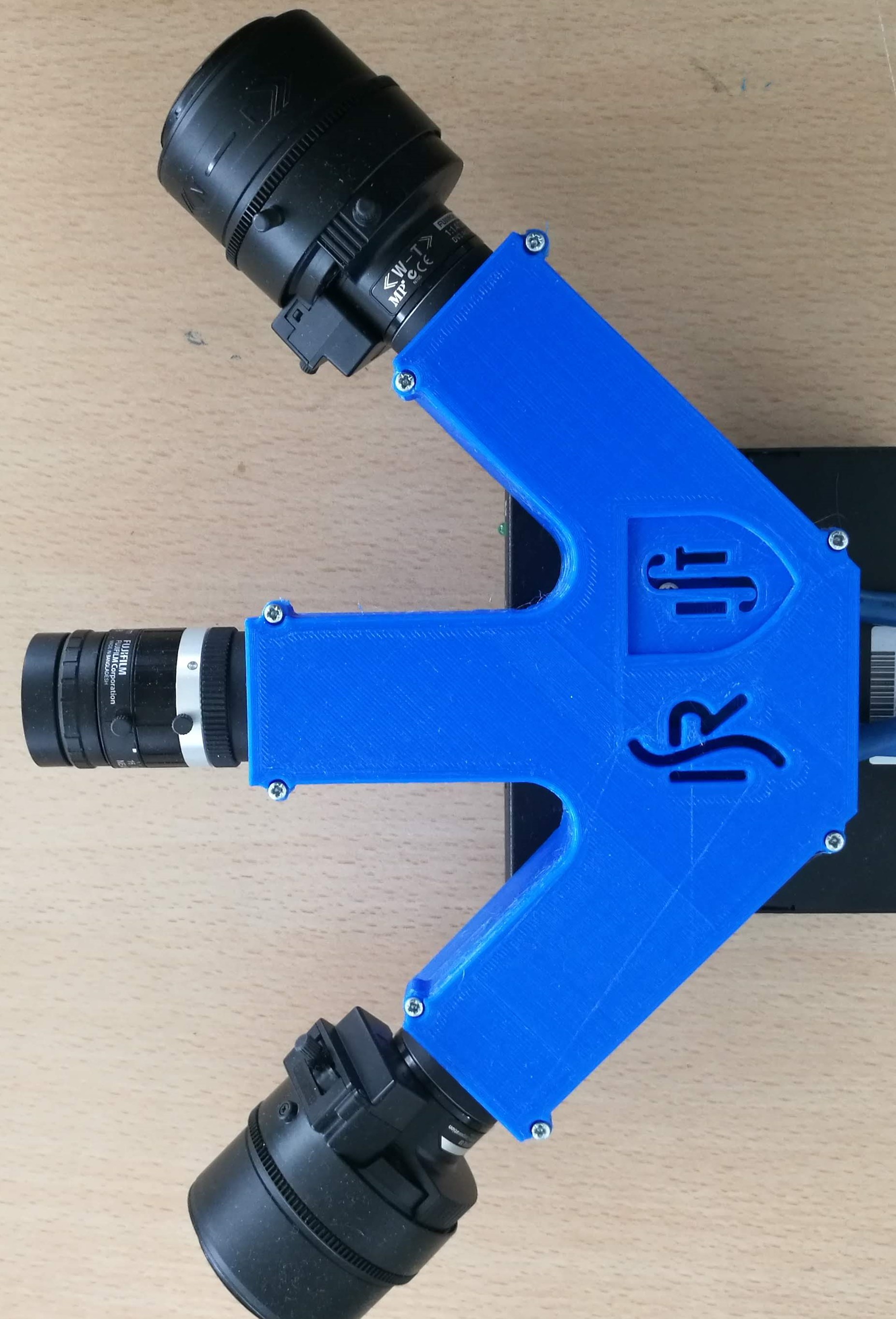}
    \label{fig:real2:setup}
    }
    \end{tabular}
    \begin{tabular}[b]{c}%
    \subfloat{\includegraphics[height=0.063\textheight]{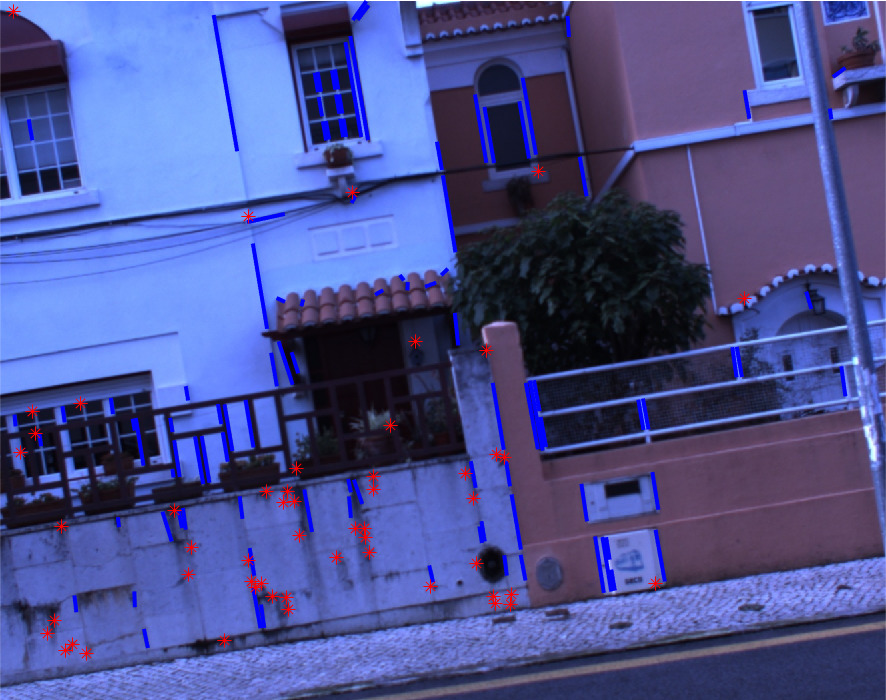}
    \includegraphics[height=0.063\textheight]{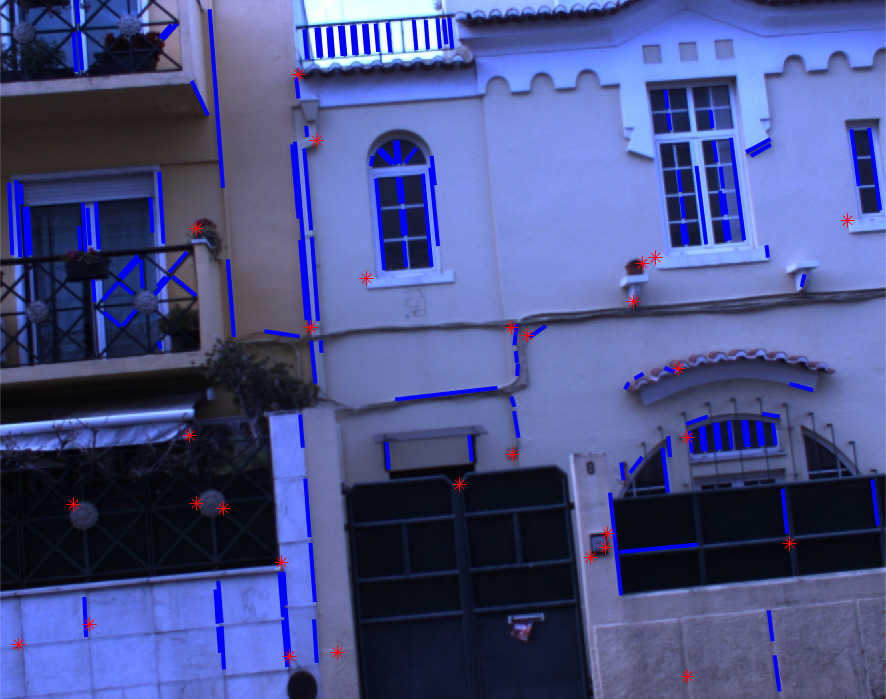}} \\
    \subfloat{\includegraphics[height=0.06\textheight]{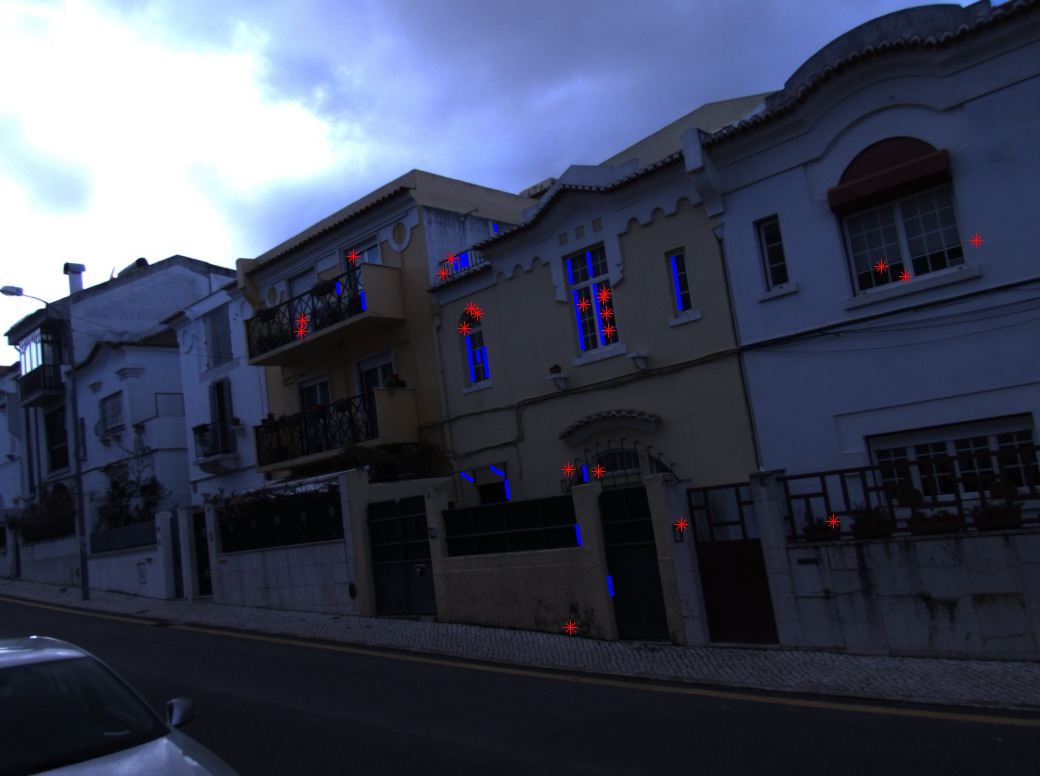}
    \includegraphics[height=0.06\textheight]{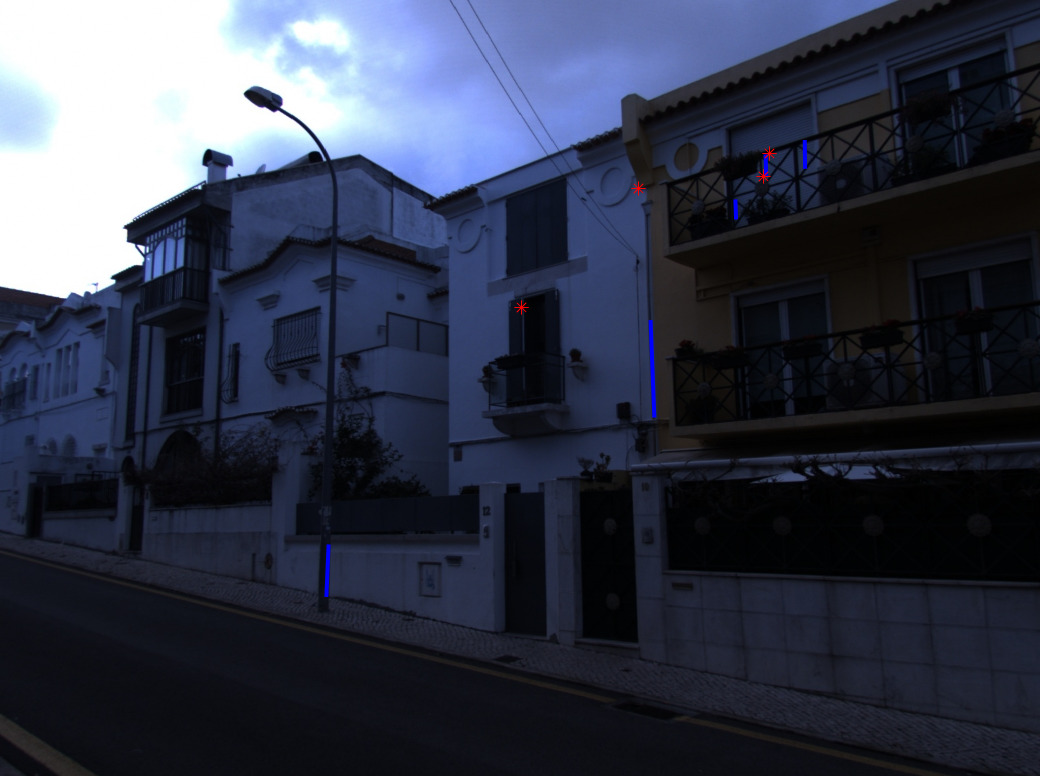}} \\
    \addtocounter{subfigure}{-2}
    \subfloat[Sample images.]{\includegraphics[height=0.06\textheight]{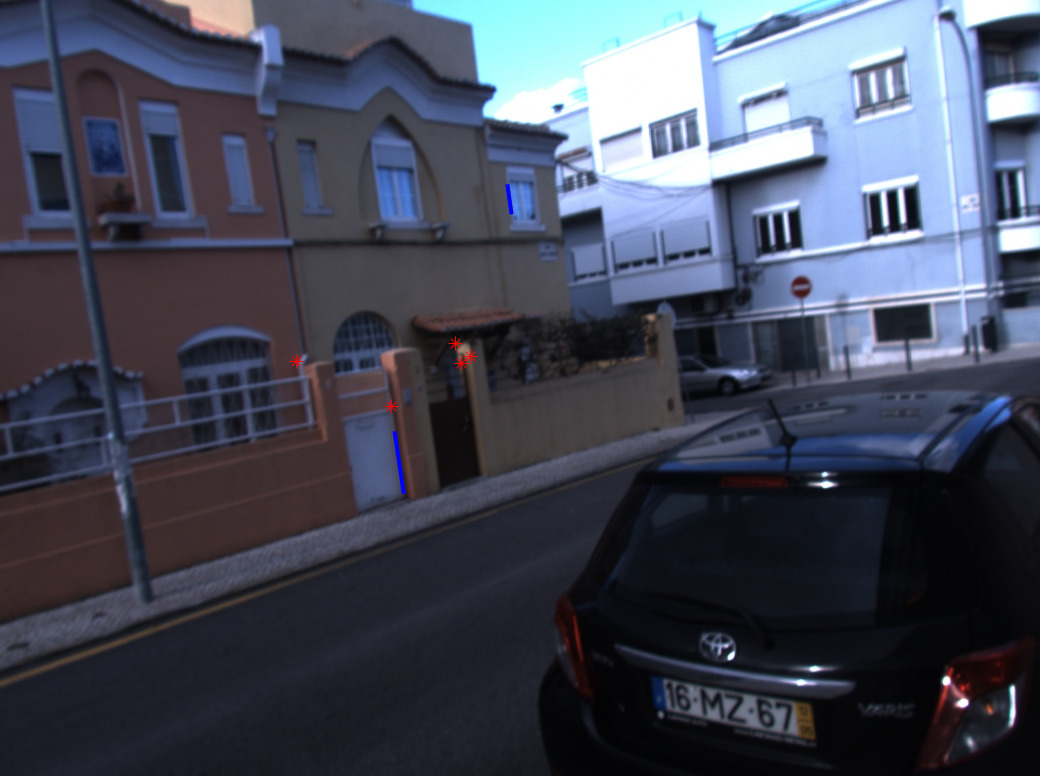}
    \includegraphics[height=0.06\textheight]{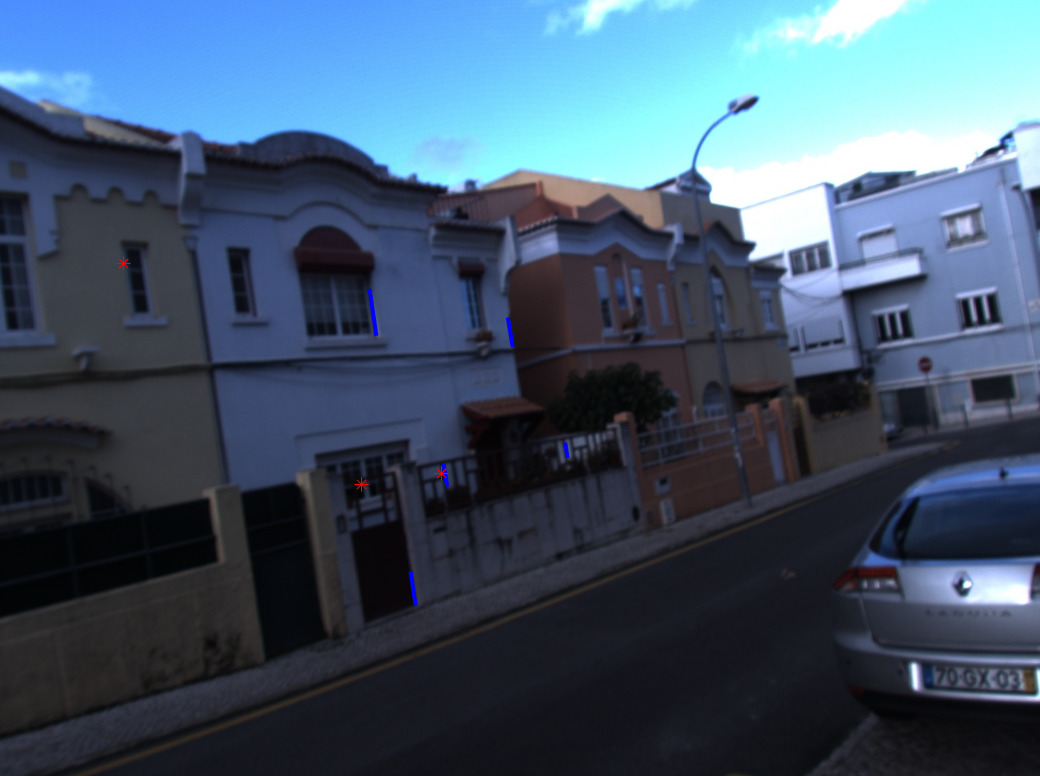}
    \label{fig:real2:images}}
    \end{tabular}
  \begin{tabular}[b]{c}%
    \subfloat[Recovered path.]{\includegraphics[height=0.194\textheight]{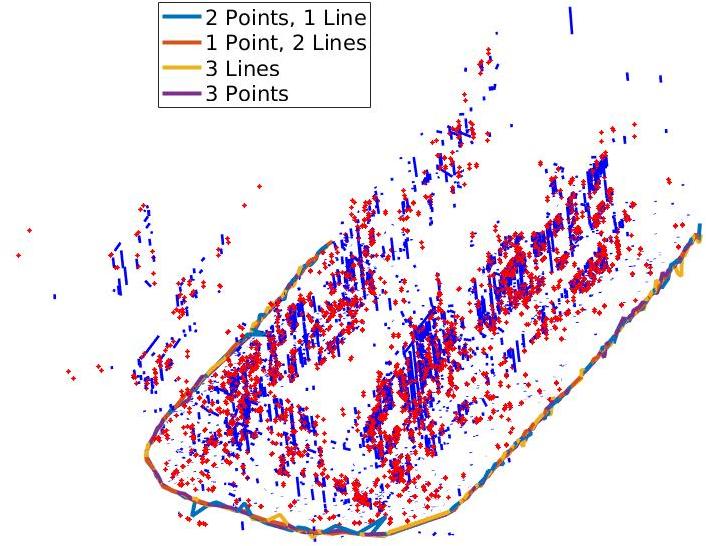}
    \label{fig:real2:results}
    }
    \end{tabular}
    }
    \caption{Results of our methods in the path estimation, using a RANSAC framework. At the left, we show the used imaging device (three cameras with angles of 45 degrees between them). In the middle, we show two columns representing two sequences acquired at the same instance by our camera system. At the right, we show a reconstructed path obtained using all methods evaluated.}
    \label{fig:real2}
\end{figure}

\subsection{Path Reconstruction using a Multi-Perspective System}
\label{sec:real}
Using the presented methods, we demonstrate a 3D reconstruction pipeline for a multi-perspective camera. For that purpose, we use a real imaging device (see Fig.~\ref{fig:real2}\subref{fig:real2:setup}), and acquired several images from an outdoor environment. We extract the correspondences between world and image features as follows:
\begin{itemize}
    \item We get camera poses and correspondences between 3D points and image pixels using the {\tt VisualSFM} framework \cite{wu11,wu13}; and
    \item To get the line correspondences, we use the {\tt Line3D++} algorithm, \cite{hofer17}. This method requires as input the camera positions, in which we use the poses given by the {\tt VisualSFM} application.
\end{itemize}

Then, we calibrate each camera individually, using the {\tt Matlab} calibration toolbox. The transformations parameters $\mathbf{T}_{\mathcal{C}_i\mathcal{C}}$ are given by the system's CAD model. Then, we run both methods proposed in this paper and existing solutions, using the RANSAC framework with a $30\%$ of required inliers and thresholds used in the previous subsection. The data-set, including images, 3D-2D correspondences (for both lines and points) and camera system calibration are available in the author's website, as well as a video with the reconstructed paths for this experiment. A total of 606 images were taken from a path of around 200 meters (examples of these pictures are shown in Fig.~\ref{fig:real2}\subref{fig:real2:images}). An average of 130 lines and 50 points per image were used, within a total of 5814 3D lines and 2230 3D points in the world.

Fig.~\ref{fig:real2}\subref{fig:real2:results} shows the results of the path reconstruction using various solvers, and they produce similar results.           
\section{Discussion}
\label{sec:conclusion}
We present 2 minimal solvers for a multi-perspective camera: (a) using 2 points and 1 line yielding 4 solutions, and (b) using 2 lines and 1 point yielding 8 solutions. While the latter case requires iterative methods, the former can be solved efficiently in closed form. To the best of our knowledge, there is no prior work on using hybrid features for a multi-camera system. Note that existing solutions (i.e. using only points or lines) require the use of iterative techniques. 

We show comparison with other minimal solvers, and we perform similar or superior to the ones that solely use points or lines. While the difference in performance among different minimal solvers can only be marginal, it is more important to note that these hybrid solvers can be beneficial and robust in noisy, dynamic, and challenging on-road scenarios where it is difficult to even get a few good correspondences. We also demonstrate a real experiment to recover the path of an outdoor sequence using a 3-camera system.   

Our method can be seen as a generalization of existing pose solvers for central cameras that uses points and lines correspondences. If we set $\mathbf{T}_{\mathcal{C}_1\mathcal{C}} = \mathbf{T}_{\mathcal{C}_2\mathcal{C}} = \mathbf{T}_{\mathcal{C}_3\mathcal{C}}$, our method solves the problem of minimal problem for perspective cameras as the current state-of-the-art method.

\section*{Acknowledgments}
P. Miraldo and T. Dias are with the Institute for Systems and Robotics (ISR/IST), LARSyS, Instituto Superior T\'{e}cnico, University of Lisboa, Portugal. This work was partially supported by the Portuguese projects [UID/EEA/50009/2013] \& [PTDC/EEI-SII/4698/2014] and grant [SFRH/BPD/111495/2015]. We thank the reviewers and ACs for valuable feedback.   

\bibliographystyle{splncs}
\bibliography{./files/egbib} 

\clearpage

\title{Supplementary Material} 

\titlerunning{Multi-Perspective Camera Pose using Points and Lines}
%
\author{Pedro Miraldo\textsuperscript{1}
\and Tiago Dias\textsuperscript{1}
\and Srikumar Ramalingam\inst{2}}
%
\authorrunning{P. Miraldo, T. Dias, and S. Ramalingam}
%

\institute{Instituto Superior T\'{e}cnico, Lisboa \\ 
\email{\{pedro.miraldo,tiagojdias\}@tecnico.ulisboa.pt}
\and
School of Computing, University of Utah\\
\email{srikumar@cs.utah.edu}}

\maketitle
\appendix
This supplementary material presents details to support the results presented in the paper: A Minimal Closed-Form Solution for Multi-Perspective Pose Estimation using Points and Lines, European Conf. on Computer Vision (ECCV'18).

\section{Getting the Predefined Transformations}
\label{sec:get_predef}

In this section we show a way of getting the predefined transformation used to obtain the data as indicated in Sec.~3.1 of the main paper\footnote{Notice that there may be alternatives that can be used to achieve the same specifications.}. Specifically, we present the rotation and translation parameters that transform the data from the world and camera coordinates, in order to verify the specifications presented in the paper.

\vspace{0.25cm}\noindent{\bf Predefined transformation to the world data:~}For a line represented using {Pl\"{u}cker} coordinates (see the notation in the paper), to have the required specifications, one has to apply the transformation:
\begin{equation}
    \tilde{\mathbf{T}}_1 =
    \begin{bmatrix}
    \tilde{\mathbf{R}}_1 & \tilde{\mathbf{t}}_1 \\
    \mathbf{0} & 1
    \end{bmatrix},
\end{equation}
such that:
\begin{multline}
    \tilde{\mathbf{R}}_1 =
    \begin{bmatrix} \mathbf{r}_1 & \mathbf{r}_2 & \mathbf{r}_3 \end{bmatrix}^T,
    \ \text{where} \\
    \mathbf{r}_1 = \frac{\bar{\mathbf{l}}_1^{\mathcal{W}}\times(\mathbf{p}_2^{\mathcal{W}}-\mathbf{q}_1^{\mathcal{W}})}{\| \bar{\mathbf{l}}_1^{\mathcal{W}}\times(\mathbf{p}_2^{\mathcal{W}}-\mathbf{q}_1^{\mathcal{W}}) \|} \ , \ \
    \mathbf{r}_2 = \frac{\bar{\mathbf{l}}_1^{\mathcal{W}}}{\| \bar{\mathbf{l}}_1^{\mathcal{W}} \|} \ \text{and} \ \
    \mathbf{r}_3 = \mathbf{r}_1 \times \mathbf{r}_2,
\end{multline}
and:
\begin{equation}
    -\tilde{\mathbf{t}}_1 = \tilde{\mathbf{R}}_1\mathbf{q}_1^{\mathcal{W}} +
    \lambda \begin{bmatrix} 0 & 1 & 0 \end{bmatrix}^T, \ \text{where} \ 
    \lambda = \left[\tilde{\mathbf{R}}_1\mathbf{p}_2^{\mathcal{W}}\right]_2 - \left[\tilde{\mathbf{R}}_1\mathbf{q}_1^{\mathcal{W}}\right]_2
\end{equation}
$\mathbf{q}_1^{\mathcal{W}}$ is any point on the line $\mathbf{l}_1^{\mathcal{W}}$ and ${[.]}_2$ denotes the second element of the vector.

\vspace{0.25cm}\noindent{\bf Predefined transformation to the camera data:~} When considering the data in the camera coordinates, we propose the following transformations:
\textbf{\begin{equation}
    \tilde{\mathbf{T}}_2 =
    \begin{bmatrix}
    \tilde{\mathbf{R}}_2 & \tilde{\mathbf{t}}_2 \\
    \mathbf{0} & 1
    \end{bmatrix},
\end{equation}}
such that:
\begin{equation}
    \tilde{\mathbf{R}}_2 =
    \begin{bmatrix} \mathbf{r}_1 & \mathbf{r}_2 & \mathbf{r}_3 \end{bmatrix}^T,
    \ \text{where} \
    \mathbf{r}_1 = \frac{\bm{e}}{\| \bm{e} \| }, \
    \mathbf{r}_2 = \ \mathbf{r}_3 \times \mathbf{r}_1 , \ \text{and} \
    \mathbf{r}_3 = \frac{\bar{\bm{\pi}}_1^{\mathcal{C}}}{\| \bar{\bm{\pi}}_1^{\mathcal{C}} \|},
\end{equation}
where $\bm{e}$ is the vector with the maximum norm within $\{ \begin{bmatrix} 1 & 0 & 0 \end{bmatrix} \times \mathbf{r}_3 \ , \  \begin{bmatrix} 0 & 1 & 0 \end{bmatrix} \times \mathbf{r}_3 \}$, and:
\begin{equation}
    \tilde{\mathbf{t}}_2 = -\tilde{\mathbf{R}}_2\mathbf{c}_1^{\mathcal{C}}.
\end{equation}

\vspace{0.25cm}\noindent{\bf Get the correct camera pose:~} After estimating the camera pose using the methods described in the paper (let denote them as $\hat{\mathbf{T}}_{\mathcal{C}\mathcal{W}}$), one needs to recover the pose without these predefined transformations. Therefore, to get the real camera pose, we apply:
\begin{equation}
    \mathbf{T}_{\mathcal{C}\mathcal{W}} = \tilde{\mathbf{T}}_1^{-1} \hat{\mathbf{T}}_{\mathcal{C}\mathcal{W}} \tilde{\mathbf{T}}_2.
\end{equation}
\section{Getting the Fourth Degree Polynomial Equation for the Two Points and One Line Solutions}

In Sec.~3.2, we derived an algorithm for the minimal absolute pose problem of a multi-perspective camera system, using two points and one line. The method requires the estimation of the intersection points between two quadratic curves, shown in Eq. 17 of the paper. In this section, we describe a closed-form method that we used to solve this problem.

Solving $\kappa_{16}^2[\delta_2,\delta_3]$ for the unknown $\delta_3$ we get two solutions of the form:
\begin{equation}
  \label{eq:unknown_b_solved}
  \delta_3 = \frac{\kappa_{20}^1[\delta_2] \pm \sqrt{\kappa_{21}^2[\delta_2]}}{\kappa_{22}^1[\delta_2]}.
\end{equation}

Substituting $\delta_3$ in $\kappa_{18}^2[\delta_2,\delta_3]$ by (\ref{eq:unknown_b_solved}), and multiplying the result by $\left.\kappa_{22}^1[\delta_2]\right.^2$ we get a constraint:
\begin{equation}
  \label{eq:unknown_c_solve_1}
  \kappa_{40}^2[\delta_2]  \pm \kappa_{41}^1[\delta_2]\sqrt{\kappa_{21}^2[\delta_2]} = 0,
\end{equation}
which we implicitly increase the degree by multiplying both solutions:
\begin{multline}
  \label{eq:unknown_c_solve_2}
  \left(\kappa_{40}^2[\delta_2]  + \kappa_{41}^1[\delta_2]\sqrt{\kappa_{21}^2[\delta_2]}\right)
  \left(\kappa_{40}^2[\delta_2]  - \kappa_{41}^1[\delta_2]\sqrt{\kappa_{21}^2[\delta_2]}\right) = 0 \Rightarrow \\
  \left.\kappa_{40}^2[\delta_2]\right.^2 - \left.\kappa_{41}^1[\delta_2]\right.^2\kappa_{21}^2[\delta_2] = 0 \Rightarrow \kappa_{19}^4[\delta_2] = 0
\end{multline}
where the polynomial equation $\kappa_{19}^4[\delta_2]$ has degree four.
\section{Coefficients of the Polynomial Equations Presented in the Paper}

This appendix presents the coefficients of the polynomial equations presented in the paper. We start by the coefficients used for the derivations of the two points and one line problem (Sec.~\ref{sec_app:2p1l_coeffs}), and then present the coefficients for one point and two lines (Sec.~\ref{sec_app:1p2l_coeffs}).

\subsection{Two points and one line coefficients}\label{sec_app:2p1l_coeffs}
In this section we present the main polynomial equations presented in the paper. We start from Eqs. 7, 8, and 9 of the paper, which represents the effects of the first collinearity constraint, which get us $t_1$, $t_2$, and $t_3$ as a function of $c\theta$, $s\theta$, $c\alpha$, $s\alpha$, $\delta_2$:
\begin{align}
    t_1 = \kappa_1^3[c\theta,s\theta,c\alpha,s\alpha,\delta_2] = &~ -d_{2,1} c\theta c2 \delta_2 - c_{2,1} c\theta c2 - d_{2,2} c\theta \delta_2 s\alpha - \nonumber \\ &~ c_{2,2} c\theta s\alpha + d_{2,3} \delta_2 s\theta + c_{2,3} s\theta,\\
    t_2 =  \kappa_2^2[c\alpha,s\alpha,\delta_2] = &~  -d_{2,2} c\alpha \delta_2 - c_{2,2}  c\alpha, -d_{2,1} \delta_2 s\alpha + c_{2,1} s\alpha, \text{ and} \\
    t_3 =  \kappa_3^3[c\theta,s\theta,c\alpha,s\alpha,\delta_2] = &~ -d_{2,3} c\theta \delta_2 - c_{2,3} c\theta - d_{2,1} c\alpha \delta_2 s\theta - c_{2,1} c\alpha s\theta - \nonumber \\ &~  d_{2,2} \delta_2 s\theta s\alpha - c_{2,2} s\theta s\alpha + p_{2,3},
\end{align}
where $p_{i,j}$, $c_{i,j}$, and $d_{i,j}$ denotes the $j$\textsuperscript{th} element of the vectors $\mathbf{p}_i^{\mathcal{W}}$, $\mathbf{c}_i^{\mathcal{W}}$, and $\mathbf{d}_i^{\mathcal{W}}$, respectively (defined in the paper).

Moving to the constraints associated with the second collinearity constraint (associated with the use of $\mathbf{p}_3^{\mathcal{W}}$), i.e. the Eqs. 11, 12, 13, and 14, we get:
\begin{align}
\kappa_4^3[c\theta,s\theta,c\alpha,s\alpha,\delta_2,\delta_3] = &~
-d_{2,1} c\theta c\alpha \delta_2 + 
d_{3,1} c\theta c\alpha \delta_3 +  \nonumber \\ &~
(c_{3,1} - c_{2,1}) c\theta c\alpha - 
d_{2,2} c\theta \delta_2 s\alpha +
d_{3,2} c\theta \delta_3 s\alpha +  \nonumber \\ &~
(c_{3,2} - c_{2,2}) c\theta s\alpha +  
d_{2,3} \delta_2 s\theta -
d_{3,3} \delta_3 s\theta + \nonumber \\ &~
(c_{2,3} - c_{3,3}) s\theta - 
p_{3,1} = 0, \\
\kappa_5^2[c\alpha,s\alpha,\delta_2,\delta_3]  = &~
-d_{2,2} c\alpha \delta_2 + 
d_{3,2} c\alpha \delta_3 +  \nonumber \\ &~
(c_{3,2} - c_{2,2}) c\alpha + 
d_{2,1} \delta_2 s\alpha -  
d_{3,1} \delta_3 s\alpha + \nonumber \\ &~
(c_{2,1} - c_{3,1}) s\alpha - 
p_{3,2} = 0, \\
\kappa_6^3[c\theta,s\theta,\delta_2,\delta_3] =  &~
-d_{2,3} c\theta \delta_2 + 
d_{3,3} c\theta \delta_3 +  \nonumber \\ &~
(c_{3,3} - c_{2,3}) c\theta -
d_{2,1} c\alpha \delta_2 s\theta + 
d_{3,1} c\alpha \delta_3 s\theta +   \nonumber \\ &~
(c_{3,1} - c_{2,1}) c\alpha s\theta -
d_{2,2} \delta_2 s\theta s\alpha +
d_{3,2} \delta_3 s\theta s\alpha + \nonumber \\ &~
(c_{3,2} - c_{2,2}) s\theta s\alpha +
(p_{2,3}-p_{3,3}) = 0, \text{ and} \\
\kappa_7^3[c\theta,s\theta,\delta_2] = &~
-d_{2,3} c\theta^2 \delta_2 -
c_{2,3} c\theta^2 +
p_{2,3} c\theta - 
d_{2,3} \delta_2 s\theta^2 - \nonumber \\ &~
c_{2,3} s\theta^2.
\end{align}

Now solving the above equations as a function of $c\theta$, $s\theta$, $c\alpha$, and $s\alpha$ and replacing these values in the trigonometric constraints $c\theta^2 + s\theta^2 = 1$ and $c\alpha^2 + s\alpha^2 = 1$, we get the solution to our problem by the following two polynomial equations:
\begin{align}
    \kappa_{16}^2[\delta_2,\delta_3] & = 
    a_1 \delta_2^2 + a_2 \delta_2 \delta_3 + a_3 \delta_2 + a_4 \delta_3^2 + a_5 \delta_3 + a_6 \text{ and}\\
    \kappa_{18}^2[\delta_2,\delta_3] & =
    a_7 \delta_2^2 + a_8 \delta_2 \delta_3 + a_9 \delta_2 + a_{10} \delta_3^2 + a_{11} \delta_3 + a_{12}, 
\end{align}
where:
\begin{align}
    a_1 & = -d_{2,3}^2 (p_{3,1}^2 + p_{3,3}^2),\\
    a_2 & = 2 d_{2,3} d_{3,3} p_{2,3} p_{3,3},\\
    a_3 & = -2 d_{2,3} (c_{2,3} p_{3,1}^2 + c_{2,3} p_{3,3}^2 - c_{3,3} p_{2,3} p_{3,3}),\\
    a_4 & = -d_{3,3}^2 p_{2,3}^2,\\
    a_5 & = 2 d_{3,3} p_{2,3} (c_{2,3} p_{3,3} - c_{3,3} p_{2,3}),\\
    a_5 & = - c_{2,3}^2 p_{3,1}^2 - c_{2,3}^2 p_{3,3}^2 + 2 c_{2,3} c_{3,3} p_{2,3} p_{3,3} - c_{3,3}^2 p_{2,3}^2 + p_{2,3}^2 p_{3,1}^2;
\end{align}
and
\begin{align}
    a_7 = &~ - d_{2,1}^2 p_{2,3}^2 p_{3,1}^2 - d_{2,2}^2 p_{2,3}^2 p_{3,1}^2 + d_{2,3}^2 p_{2,3}^2 p_{3,3}^2 - \nonumber \\ &~ 2 d_{2,3}^2 p_{2,3} p_{3,1}^2 p_{3,3} - 2 d_{2,3}^2 p_{2,3} p_{3,3}^3 + d_{2,3}^2 p_{3,1}^4 + 2 d_{2,3}^2 p_{3,1}^2 p_{3,3}^2 + d_{2,3}^2 p_{3,3}^4,\\
    a_8 = &~ 2 d_{2,1} d_{3,1} p_{2,3}^2 p_{3,1}^2 - 2 d_{2,3} d_{3,3} p_{2,3}^3 p_{3,3} - 2 d_{2,3} d_{3,3} p_{2,3} p_{3,3}^3 + \nonumber \\ &~ 2 d_{2,2} d_{3,2} p_{2,3}^2 p_{3,1}^2 + 2 d_{2,3} d_{3,3} p_{2,3}^2 p_{3,1}^2 + 4 d_{2,3} d_{3,3} p_{2,3}^2 p_{3,3}^2 -  \nonumber \\ &~ 2 d_{2,3} d_{3,3} p_{2,3} p_{3,1}^2 p_{3,3},\\
    a_9 = &~ 2 c_{2,3} d_{2,3} p_{3,1}^4 + 2 c_{2,3} d_{2,3} p_{3,3}^4 - 4 c_{2,3} d_{2,3} p_{2,3} p_{3,3}^3 - 2 c_{3,3} d_{2,3} p_{2,3} p_{3,3}^3 -  \nonumber \\ &~ 2 c_{3,3} d_{2,3} p_{2,3}^3 p_{3,3} - 2 c_{2,1} d_{2,1} p_{2,3}^2 p_{3,1}^2 - 2 c_{2,2} d_{2,2} p_{2,3}^2 p_{3,1}^2 +  \nonumber \\ &~ 2 c_{2,3} d_{2,3} p_{2,3}^2 p_{3,3}^2 + 2 c_{3,1} d_{2,1} p_{2,3}^2 p_{3,1}^2 + 2 c_{3,2} d_{2,2} p_{2,3}^2 p_{3,1}^2 +  \nonumber \\ &~ 4 c_{2,3} d_{2,3} p_{3,1}^2 p_{3,3}^2 + 2 c_{3,3} d_{2,3} p_{2,3}^2 p_{3,1}^2 + 4 c_{3,3} d_{2,3} p_{2,3}^2 p_{3,3}^2 -  \nonumber \\ &~ 4 c_{2,3} d_{2,3} p_{2,3} p_{3,1}^2 p_{3,3} - 2 c_{3,3} d_{2,3} p_{2,3} p_{3,1}^2 p_{3,3},\\
    a_{10} = &~ -p_{2,3}^2 (d_{3,1}^2 p_{3,1}^2 + d_{3,2}^2 p_{3,1}^2 - d_{3,3}^2 p_{2,3}^2 + 2 d_{3,3}^2 p_{2,3} p_{3,3} - d_{3,3}^2 p_{3,3}^2),\\
    a_{11} = &~ 2 c_{3,3} d_{3,3} p_{2,3}^4 - 2 c_{2,3} d_{3,3} p_{2,3} p_{3,3}^3 - 2 c_{2,3} d_{3,3} p_{2,3}^3 p_{3,3} -  \nonumber \\ &~ 4 c_{3,3} d_{3,3} p_{2,3}^3 p_{3,3} + 2 c_{2,1} d_{3,1} p_{2,3}^2 p_{3,1}^2 + 2 c_{2,2} d_{3,2} p_{2,3}^2 p_{3,1}^2 +  \nonumber \\ &~ 2 c_{2,3} d_{3,3} p_{2,3}^2 p_{3,1}^2 + 4 c_{2,3} d_{3,3} p_{2,3}^2 p_{3,3}^2 - 2 c_{3,1} d_{3,1} p_{2,3}^2 p_{3,1}^2 -  \nonumber \\ &~ 2 c_{3,2} d_{3,2} p_{2,3}^2 p_{3,1}^2 + 2 c_{3,3} d_{3,3} p_{2,3}^2 p_{3,3}^2 - 2 c_{2,3} d_{3,3} p_{2,3} p_{3,1}^2 p_{3,3},\\
    a_{12} = &~ - c_{2,1}^2 p_{2,3}^2 p_{3,1}^2 + 2 c_{2,1} c_{3,1} p_{2,3}^2 p_{3,1}^2 - c_{2,2}^2 p_{2,3}^2 p_{3,1}^2 + 2 c_{2,2} c_{3,2} p_{2,3}^2 p_{3,1}^2 +  \nonumber \\ &~ c_{2,3}^2 p_{2,3}^2 p_{3,3}^2 - 2 c_{2,3}^2 p_{2,3} p_{3,1}^2 p_{3,3} - 2 c_{2,3}^2 p_{2,3} p_{3,3}^3 + c_{2,3}^2 p_{3,1}^4 +   \nonumber \\ &~ 2 c_{2,3}^2 p_{3,1}^2 p_{3,3}^2 + c_{2,3}^2 p_{3,3}^4 - 2 c_{2,3} c_{3,3} p_{2,3}^3 p_{3,3} + 2 c_{2,3} c_{3,3} p_{2,3}^2 p_{3,1}^2 +   \nonumber \\ &~ 4 c_{2,3} c_{3,3} p_{2,3}^2 p_{3,3}^2 - 2 c_{2,3} c_{3,3} p_{2,3} p_{3,1}^2 p_{3,3} - 2 c_{2,3} c_{3,3} p_{2,3} p_{3,3}^3 -   \nonumber \\ &~ c_{3,1}^2 p_{2,3}^2 p_{3,1}^2 - c_{3,2}^2 p_{2,3}^2 p_{3,1}^2 + c_{3,3}^2 p_{2,3}^4 - 2 c_{3,3}^2 p_{2,3}^3 p_{3,3} +   \nonumber \\ &~ c_{3,3}^2 p_{2,3}^2 p_{3,3}^2 + p_{2,3}^2 p_{3,1}^2 p_{3,2}^2.
\end{align}

\subsection{One point and two lines coefficients}\label{sec_app:1p2l_coeffs}
Now, let us consider the coefficients of the polynomial equations for the one point and two lines case. We start from Eq. 20 of the paper in which, from Eq. 14 of the paper and the trigonometric relation $c\theta^2 + s\theta^2 = 1$, gets a relation between $c\theta$ and $s\theta$ as a function of the unknown $\delta_2$:
\begin{align}
    \kappa_{23}^1[\delta_2] = &~ \frac{c_{2,3} + d_{2,3} \delta_2}{p_{2,3}}, \text{ and} \\
    \kappa_{24}^2[\delta_2] = &~ \frac{(c_{2,3} + p_{2,3} + d_{2,3} \delta_2) (c_{2,3} - p_{2,3} + d_{2,3} \delta_2)}{p_{2,3}^2}.
    \label{eq:sol_stheta}
\end{align}

Next, we use the same collinearity constraint derived in Eqs. 7, 8, and 9 of the paper, and the coplanarity constraints (Eq. 21) to find $c\alpha$ and $s\alpha$ as a function of $\delta_2$ and $s\theta$, which is shown in Eq. 23 of the paper. Then, we use this result in the trigonometric equation $c\alpha^2 + s\alpha^2 = 1$, and get a single constraint as a function of $\delta_2$ and $s\theta$:
\begin{multline}
    \kappa_{29}^4[s\theta,\delta_2] = b_1 s\theta^4 + b_2 s\theta^3 \delta_2 + b_3 s\theta^3 + b_4 s\theta^2 \delta_2^2 + b_5 s\theta^2 \delta_2 + b_6 s\theta^2 + \\ b_7 s\theta \delta_2^3 + b_8 s\theta \delta_2^2 + b_9 s\theta \delta_2 + b_{10} s\theta + b_{11} \delta_2^4 + b_{12} \delta_2^3 + b_{13} \delta_2^2 + b_{14} \delta_2 + b_{15},
\end{multline}
the parameters $b_1$ \dots $b_{15}$ are sent in the {\tt polynomial.txt} file. In order to get the $s\theta$ with even power numbering, we get the following polynomial equations presented in Eq. 25 of the paper:
\begin{align}
    \kappa_{35}^2[\delta_2] = &~ b_{16}\delta_2^2 + \delta_2b_{17} + b_{18}\\
    \kappa_{36}^4[\delta_2] = &~ b_{19}\delta_2^4 + b_{20}\delta_2^3 + b_{21}\delta_2^2 + b_{22}\delta_2 + b_{23}\\
    \kappa_{37}^6[\delta_2] = &~ b_{24}\delta_2^6 + b_{25}\delta_2^5 + b_{26}\delta_2^4 + b_{27}\delta_2^3 + b_{28}\delta_2^2 + b_{29}\delta_2 + b_{30}\\
    \kappa_{38}^8[\delta_2] = &~ b_{31}\delta_2^8 + b_{32}\delta_2^7  + b_{33}\delta_2^6 + b_{34}\delta_2^5 + b_{35}\delta_2^4 + b_{36}\delta_2^3 + b_{37}\delta_2^2 + b_{38}\delta_2 + b_{39}
\end{align}

To conclude, substituting the $s\theta^2$ by \eqref{eq:sol_stheta}, we get the final eight degree polynomial equation:
\begin{equation}
    \kappa_{38}^8[\delta_2] =  f_{1}\delta_2^8 + f_{2}\delta_2^7  + f_{3}\delta_2^6 + f_{4}\delta_2^5 + f_{5}\delta_2^4 + f_{36}\delta_2^6 + f_{7}\delta_2^2 + f_{8}\delta_2 + f_{9}.
\end{equation}
The final coefficients $f_i$ for $i=1,\dots 9$ are shown in the file  {\tt polynomial.txt}.

\end{document}